\documentclass[review]{elsarticle}

\usepackage{hyperref}

\usepackage{microtype}
\usepackage{graphicx}
\usepackage{booktabs} 
\usepackage[utf8]{inputenc}
\usepackage[english]{babel}
\usepackage{times}
\usepackage{epsfig}
\usepackage{amsmath}
\usepackage{amsthm}
\usepackage{color}
\usepackage{xcolor}
\usepackage{amssymb}
\usepackage{multirow}
\usepackage[normalem]{ulem}
\usepackage[caption=false,font=footnotesize]{subfig}
\usepackage{amsmath,amssymb} 
\usepackage[linesnumbered]{algorithm2e}
\usepackage{float}
\usepackage{tikz}

\SetKwInput{KwInput}{Input}                
\SetKwInput{KwOutput}{Output}              

\newtheorem{theorem}{Theorem}[section]

\newtheorem{lemma}[theorem]{Lemma}

\journal{Pattern Recognition}

\begin{document}

\begin{frontmatter}

\title{LongReMix: Robust Learning with High Confidence Samples in a Noisy Label Environment}

\author[ufrpe, pa]{Filipe R. Cordeiro\corref{mycorrespondingauthor}}
\cortext[mycorrespondingauthor]{Corresponding author}
\ead{filipe.rolim@ufrpe.br}

\author[vgg]{Ragav Sachdeva}
\author[Magdeburg]{Vasileios Belagiannis}
\author[adelaide]{Ian Reid}
\author[adelaide,surrey]{Gustavo Carneiro}

\address[adelaide]{School of Computer Science, Australian Institute for Machine Learning, Australia}
\address[vgg]{Visual Geometry Group, Department of Engineering Science, University of Oxford, United Kingdom}
\address[ufrpe]{Visual Computing Lab, Department of Computing, Universidade Federal Rural de Pernambuco, Brazil}
\address[Magdeburg]{Otto-von-Guericke-Universit\"at Magdeburg, Germany}
\address[surrey]{Centre for Vision, Speech and Signal Processing, University of Surrey, United Kingdom}

\begin{abstract}
   State-of-the-art noisy-label learning algorithms rely on an unsupervised learning to classify training samples as clean or noisy, followed by a semi-supervised learning (SSL) that minimises the empirical vicinal risk using a labelled set formed by samples classified as clean, and an unlabelled set with samples classified as noisy. 
   The classification accuracy of such noisy-label learning methods depends on the precision of the unsupervised classification of clean and noisy samples, and the robustness of SSL to small clean sets. 
   We address these points with a new noisy-label training algorithm, called LongReMix, which improves the precision of the unsupervised classification of clean and noisy samples and the robustness of SSL to small clean sets with a two-stage learning process. The stage one of LongReMix finds a small but precise high-confidence clean set, and stage two augments this high-confidence clean set with new clean samples and oversamples the clean data to increase the robustness of SSL to small clean sets.
   We test LongReMix on  CIFAR-10 and CIFAR-100 with introduced synthetic noisy labels, and the real-world noisy-label benchmarks  CNWL (Red Mini-ImageNet), WebVision, Clothing1M, and Food101-N. The results show that our LongReMix produces  significantly better classification accuracy than competing approaches, particularly in high  noise rate problems. Furthermore, our approach achieves state-of-the-art performance in most datasets. The code is available at https://github.com/filipe-research/LongReMix. 
   
\end{abstract}


\begin{graphicalabstract}
\includegraphics[width=\linewidth]{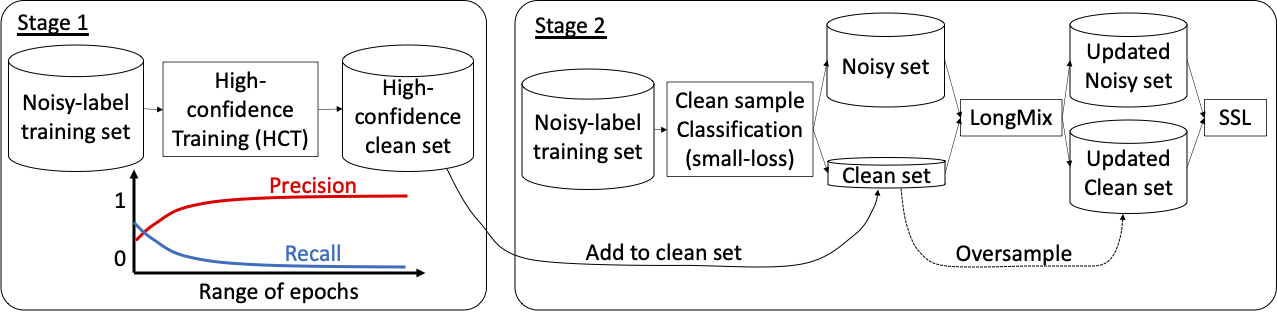}
\end{graphicalabstract}

\begin{highlights}
\item We propose a new two-stage noisy-label learning algorithm, called LongReMix;
 
\item The first stage finds a highly precise, but potentially small, set of clean samples;

\item The second stage is designed to be robust to small sets of clean samples;

\item LongReMix reaches SOTA performance on the main noisy-label learning benchmarks;

\end{highlights}

\end{frontmatter}


\section{Introduction}\label{sec:introduction}

Training Deep Neural Networks (DNNs) often requires large data sets to perform well on challenging problems such as image classification~\cite{litjens2017survey}. However, the larger the data set, the greater the likelihood for it to be contaminated with noisy labels due to reasons such as low-quality data, human failure, or challenging labelling tasks~\cite{frenay_survey}. 
The main issue is that DNNs can overfit noisy-label samples, particularly for large rate of label noise, reducing their accuracies, as shown by Zhang et al.~\cite{zhang2016understanding}. Hence the field is focusing on the development of training strategies that reduce the likelihood of DNNs to fit such noisy-label samples.

In the literature, several methods have been proposed to deal with noisy labels~\cite{kim2019nlnl, wang2019symmetric, ren2018learning,wang2019symmetric,nguyen2019self,li2020dividemix}, where one of the most successful methods explores a method formed by an unsupervised learning method to classify training samples as clean or noisy, followed by semi-supervised learning (SSL) to minimise the empirical vicinal risk (EVR) with a labelled set formed by the samples classified as clean, and an unlabelled set with the samples classified as noisy.
The unsupervised learning stage generally is based on the small-loss strategy~\cite{yu2019does}, where at every epoch, samples with small loss are classified as clean, and large loss as noisy. 
This strategy can lead to a low classification precision of clean samples, particularly in high noise rate scenarios, because the loss values can be unstable at different training epochs. The SSL stage~\cite{arazo2019unsupervised, nguyen2019self,li2020dividemix} is usually based on MixMatch~\cite{berthelot2019mixmatch} that minimises the empirical vicinal risk (EVR)~\cite{zhang2017mixup}, where a robust estimation of the vicinal distribution is critical for an effective optimisation that generalises well. 
In turn, such robust estimation depends on a large training set to minimise the EVR~\cite{berthelot2019mixmatch,zhang2018generalization}, but problems with high noise rate usually cause the unsupervised learning stage to build a small training set to be used by this optimisation, affecting the generalisation of the SSL stage.

\begin{figure}[!t]
\centering
\includegraphics[width=0.4\textwidth]{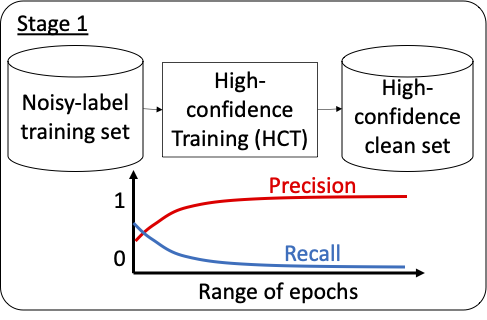}
\caption{LongReMix stage 1 finds a high-confidence set of clean samples using the small loss strategy over a range of epochs, where a larger range implies a larger precision and  lower recall in finding clean samples.}
\label{fig:motivation_part1}
\end{figure}

\begin{figure}[!t]
\centering
\includegraphics[width=0.7\textwidth]{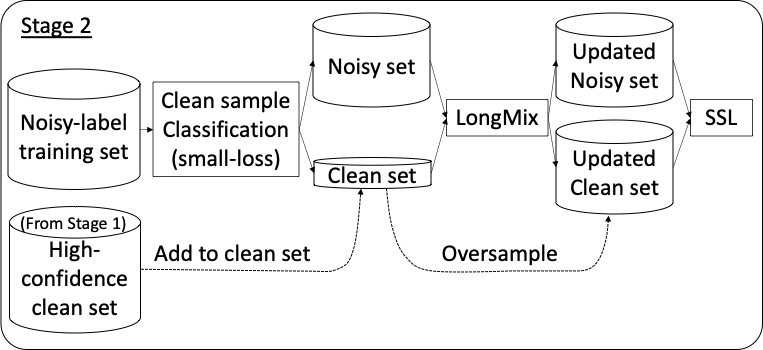}
\caption{LongReMix stage 2 increases the high-confidence clean set from stage 1 with a clean set found by the small-loss strategy, which are used to train the proposed LongMix that oversamples this clean set to improve the robustness of the SSL to small clean sets, usually formed in high label noise rate problems.}
\label{fig:motivation_part2}
\end{figure}


In this paper, we hypothesise that the classification accuracy of noisy-label learning methods depends on the precision of the unsupervised learning stage to classify clean or noisy samples and the robustness of SSL to small clean sets, formed in high label noise rate problems. 
To validate these two hypotheses, we propose LongReMix, which is a new two-stage noisy-label training algorithm. The first stage forms a high-confidence set  
of clean samples that is estimated with a new unsupervised learning method that trades off the precision and recall of this clean set, as depicted in 
Fig.~\ref{fig:motivation_part1}. 
The second stage increases the high-confidence clean set from stage 1 with a clean set estimated from the small-loss strategy, which is used to train the proposed LongMix that oversamples this clean set to improve the robustness of SSL to small clean sets, usually formed in high label noise rate problems -- see Fig.~\ref{fig:motivation_part2}.

The key contributions of LongReMix are:
\begin{itemize}
  \item A new two-stage noisy-label learning algorithm based on a highly precise unsupervised learning method to classify training samples as clean or noisy, followed by a semi-supervised learning (SSL) approach that is quite robust to small sets of clean samples.
  \item The highly precise unsupervised learning in the first stage forms a high-confidence set of clean samples using the losses of many consecutive epochs, instead of one epoch, to differentiate between noisy and clean samples; and the set of clean samples for the second stage is formed by a combination of this high-confidence clean samples (from the first stage) and the clean samples found using the the small-loss strategy (from the second stage), increasing the size and reliability of the clean set;
  \item The proposed SSL learning in the second stage estimates the vicinal distribution 
  by oversampling the clean data to increase the training set size and improve
  the robustness of EVR minimisation to small clean sets, typically formed in large label noise rate scenarios.
\end{itemize}
Even though other papers have shown that the detection of clean samples is more precise with the use of loss measures across several epochs~\cite{relab,area_under_margin,toneva2018empirical}, they tend to be unstable during early training stages, and in high-noise rate problems, they form rather small sets of clean samples that can deteriorate training robustness.
Hence, in this paper we address these two problems:
\begin{itemize}
    \item training instability at the beginning of the training, and
    \item robustness to large noise rate problems.
\end{itemize}

\noindent These two problems are addressed with the  high-confidence set (from first stage) that is combined with the small-loss strategy to form large sets of clean samples in the second stage.
We show that our high-confidence set of clean samples can be up to 30\% more precise than the clean sets formed with the small loss strategy.
Furthermore, oversampling the clean set has been explored in SSL~\cite{tarvainen2017mean, chen2018semi,iscen2019label,arazo2020pseudo}, but to the best of our knowledge, we are the first group to use it in noisy-label learning problems.

We evaluate our approach on the noisy-label learning benchmarks of CIFAR-10~\cite{krizhevsky2009learning}, CIFAR-100~\cite{krizhevsky2009learning},
Controlled Noisy Web Labels (CNWL - Red Mini-ImageNet)~\cite{jiang2020beyond},
WebVision~\cite{li2017WebVision}, Clothing1M~\cite{xiao2015learning}, and Food101-N~\cite{lee2018cleannet}, where LongReMix shows the best performance in the field in almost all of those data sets, particularly in problems with extremely large noise rates. 
These results are shown to be significantly better than  competing SOTA methods using the statistical test in~\cite{demvsar2006statistical}.
We also show that LongReMix finds a set of clean samples with higher precision than the competing methods, and is robust to over-fitting in problems with high label noise.

\section{Prior Work}

Several methods have been proposed for the noisy-label problem, and they explore different strategies, such as robust loss functions~\cite{ma2020normalized,wang2019imae, wang2019symmetric}, label cleansing \cite{jaehwan2019photometric, yuan2018iterative}, sample weighting \cite{ren2018learning}, meta-learning \cite{han2018pumpout,sun2021learning}, ensemble learning \cite{miao2015rboost}, and others \cite{yu2018learning, kim2019nlnl}. Below, we focus on the prior work that is related to our approach and show competitive results on the main benchmarks.

Sample noise characterisation can be achieved with an auxiliary clean validation set.  For instance, Ren et al.~\cite{ren2018learning} use such clean validation set and a meta-learning method to find the noisy samples and re-weight them based on their values and gradient directions.  
Zhang et al.~\cite{zhang2020distilling} also use a clean validation set and meta-learning to incorporate pseudo labeling
into meta optimisation.
Shu et al.~\cite{shu2019meta} rely on meta-learning and a clean validation set, but they use a multi-layer perceptron to learn a loss-weighting function. 
Even though these methods show competitive results, 
they need an auxiliary clean validation set that is not always available and can be expensive to acquire.
Hence, we disregard these methods in our paper because we argue that they are based on a less general experimental setup. 

The automatic characterisation of sample noise without a clean validation set has also been investigated.
Xue et al.~\cite{xue2019robust} present a probabilistic Local Outlier Factor algorithm (pLOF) to estimate the probability that a sample is an outlier, which is assumed to have a noisy label. The idea explored by pLOF is that the density around a noisy sample is significantly different from the density around its (clean) neighbors. However, in high noise rate problems, the effectiveness of pLOF is reduced because it cannot find significant differences between the densities of noisy and clean samples. 
Wang et al.~\cite{wang2018iterative} also use pLOF combined with a Siamese network to increase the dissimilarities between clean and noisy samples. Nevertheless, the incorrect classification of clean samples by pLOF can induce the learning of wrong feature representations. 
Arazo et al.~\cite{arazo2019unsupervised} propose the use of a Beta Mixture Model (BMM) to separate the clean and noise samples during training, based on the classification loss value of each sample. 
Similarly, Li et al.~\cite{li2020dividemix} use Gaussian Mixture Model (GMM) for the same goal.
Although the use of BMM and GMM applied on the loss values works well for low noise rate, for higher noise regimes it becomes less precise because these loss values are more unstable over different training epochs.  
To mitigate this vulnerability, some approaches use stored information across epochs, such as average loss~\cite{relab}, difference between logits~\cite{area_under_margin} and the number of times a sample is forgotten during training~\cite{toneva2018empirical}. 
Even though using information across epochs improves the precision of the identification of clean samples, these models are not effective in the beginning of the training, when such identification is still unstable.
Furthermore, in high-noise rate problems, such more constrained identification of clean samples can lead to small labelled training sets, deteriorating training generalisation.




We  address  these  two  problems  with  our  two-stage  training process 
(see Figures~\ref{fig:motivation_part1} and~\ref{fig:motivation_part2}),
where  the  first  stage  finds  the  
high-confidence 
set  of  labelled clean samples, and in the second stage we train the model using that 
high-confidence
labelled clean samples. More specifically, in the first stage, we classify samples into clean or noisy using their loss values over a range of training epochs. Then in the second stage, we select the largest set of labelled clean samples (from the first stage) to form a high confidence set of clean samples to be used during the rest of the training that also relies on clean samples found by the small-loss strategy. The use of this high confidence set of clean samples addresses the two issues above (i.e., training instability at the beginning of the training and robustness to large noise rate problems). To the best of our knowledge, this approach has not been explored in the field.

Another technique being studied for noisy-label learning is the use of multiple models to improve the robustness of sample noise characterisation. Han et al.~\cite{han2018co} propose Co-teaching, which trains two models simultaneously, where each model estimates the clean sample set to be used by the other model. However, with an increase in the number of epochs, both networks converge to a consensus and show little difference between their estimated clean sets. Co-teaching+~\cite{yu2019does} relies on small loss samples that disagree on the predictions to select the data for the other model. Although this multiple model strategy shows better results for filtering clean samples, noisy samples are usually ignored during training, decreasing the effectiveness of the approach. We also rely on the use of multiple models during training, but we do not ignore the noisy samples during training.

As mentioned above, after the automatic classification of clean and noisy samples, methods either disregard the noisy samples during training~\cite{thulasidasan2019combating, han2018co}, or use both the clean and noisy samples in a semi-supervised learning (SSL) approach~\cite{li2020dividemix, arazo2019unsupervised, sachdeva2021evidentialmix}, where SSL-based methods tend to show competitive results on benchmarks.
One particularly successful technique that relies on SSL is DivideMix~\cite{li2020dividemix} that relies on MixMatch~\cite{berthelot2019mixmatch} to linearly combine training samples classified as clean or noisy for the EVR minimisation~\cite{zhang2017mixup}.
The generalisation of the EVR minimisation has been theoretically shown to depend on a large training set~\cite{zhang2018generalization}.
However, DivideMix~\cite{li2020dividemix}
constrains this training set to be of the same size as the clean set, which tends to be small in large noise rate scenarios, resulting in poor EVR generalisation.
To improve the EVR generalisation in noisy-label learning problems, we propose the over-sampling of the clean set to make it as large as the original training set, containing clean- and noisy-label samples.
This idea has recently been explored in semi-supervised learning with clean labels~\cite{tarvainen2017mean, chen2018semi,iscen2019label,arazo2020pseudo}, but we are not aware of its extension to noisy-label learning problems.

\section{Problem Setup and Hypotheses}
\label{sec:problem_definition}

\begin{table}[ht]
\centering
\scriptsize
\begin{tabular}{|c|l|}
\hline
\multicolumn{1}{|c|}{Notation}& \multicolumn{1}{c|}{Description}\\
\hline\hline
$\mathcal{S}$ & image space \\
$\mathcal{Y}$ & label space \\
$\mathbf{x}$ &  image example from $\mathcal{S}$ \\
$\mathbf{y}$ &  noisy label example from $\mathcal{Y}$ \\
$\hat{\mathbf{y}}$ & true (clean) label example from $\mathcal{Y}$ \\
$\mathcal{D}=\{(\mathbf{x}_i, \mathbf{y}_i)\}_{i=1}^{|\mathcal{D}|}$ & training set \\
$\eta_{jc}(\mathbf{x}_i)$ & the probability that the label for $\mathbf{x}_i$ flips from class $c$ to $j$, ($c,j \in \{1,...,|\mathcal{Y}|\}$)\\
$f:\mathcal{S} \times \Theta \to [0,1]^{|\mathcal{Y}|}$ & classifier parameterised by $\theta \in \Theta$\\
$\Theta$ & parameter space of the classifier\\
$\ell(f(\mathbf{x};\theta),\mathbf{y})$ & classification loss to train the classifier \\
$\mathcal{X},\mathcal{U}$ & initial sets of clean and noisy samples, respectively \\
$p \left ( \text{clean} | \ell( f(\mathbf{x};\theta),\mathbf{y}) , \gamma \right )$ & probability function, parameterised by $\gamma$, that estimates if $(\mathbf{x},\mathbf{y})$ is clean \\
$\mathcal{X}'$,$\mathcal{U}'$ & clean and noisy sets formed by MixMatch~\cite{berthelot2019mixmatch} from $\mathcal{X}$,$\mathcal{U}$, respectively\\
$\ell^{(\mathcal{X}')}$, $\ell^{(\mathcal{U}')}$ & loss functions for the clean and noisy sets \\
$\ell_{EVR}$ & empirical vicinal risk that sums $\ell^{(\mathcal{X}')}$ and $\ell^{(\mathcal{U}')}$ \\
$\ell_{R}$ & regularisation loss that penalizes classifications far from the uniform distribution \\
$v(\tilde{\mathbf{x}},\tilde{\mathbf{y}}|\mathbf{x}_i,\mathbf{y}_i)$ & vicinity distribution for $(\mathbf{x}_i,\mathbf{y}_i)$ used by LongMix\\
$P_{cc}^{(e)}$ & probability of classifying a clean sample as clean during training epoch $e \in \{1,...,E\}$\\
$P_{nc}^{(e)}$ & probability of classifying a clean sample as noisy during training epoch $e \in \{1,...,E\}$\\
$P_{nn}^{(e)}$ & probability of classifying a noisy sample as noisy during training epoch $e \in \{1,...,E\}$\\
$P_{cn}^{(e)}$ & probability of classifying a noisy sample as clean during training epoch $e \in \{1,...,E\}$\\
$P_c,P_n$ &  proportion of clean and noisy
samples in the training set \\
$\mathcal{X}_1^{(e)},\mathcal{U}_1^{(e)}$ & clean and noisy sets formed by high confidence training (HCT) in the $1^{st}$ training stage\\
$\zeta$ & number of training epochs used by HCT to form $\mathcal{X}_1^{(e)},\mathcal{U}_1^{(e)}$ \\
$\mathcal{H}$ & high confidence set of clean samples estimated from $\mathcal{X}_1^{(e)}$ \\
$\mathcal{X}_2^{(e)},\mathcal{U}_2^{(e)}$ & clean and noisy sets formed in the $2^{nd}$ training stage\\
\hline
\end{tabular}
\caption{Mathematical notation.}
\label{tab:mathematical_notation}
\end{table}

We first define the noisy label learning problem. We summarise all mathematical notation used in this paper in Table~\ref{tab:mathematical_notation}.
Consider the training set $\mathcal{D}=\{(\mathbf{x}_i, \mathbf{y}_i)\}_{i=1}^{|\mathcal{D}|}$, where $\mathbf{x}_i \in \mathcal{S} \subset \mathbb{R}^{W \times H}$ is the $i^{th}$ image ($W \times H$ denote the width and height of the image) and $\mathbf{y}_i \in \{0,1\}^{|\mathcal{Y}|}$ is a one-hot vector representing the noisy label, with $\mathcal{Y} = \{1,...,|\mathcal{Y}|\}$ denoting the set of labels, and $\sum_{c \in \mathcal{Y}} \mathbf{y}_i(c)=1$. 
The label $\mathbf{y}_i$ may differ from the unknown true label $\hat{\mathbf{y}}_i$ as a result of a noise process represented by $\mathbf{y}_i \sim p(\mathbf{y} | \mathbf{x}_i,\mathcal{Y},\hat{\mathbf{y}}_i)$, with $p(\mathbf{y}(j) | \mathbf{x}_i,\mathcal{Y},\hat{\mathbf{y}}_i(c))=\eta_{jc}(\mathbf{x}_i)$,
where the $j,c\in\{1,...,\mathcal{Y}\}$ are the class indexes, $\eta_{jc}(\mathbf{x}_i) \in [0,1]$ the probability of flipping from class $c$ to $j$, and $\sum_{j \in \mathcal{Y}}\eta_{jc}(\mathbf{x}_i)=1$. We assume that this noise process can be of three types, namely symmetric~\cite{kim2019nlnl}, asymmetric~\cite{patrini2017making}, and instanced-based~\cite{rog}.
The symmetric noise, also called uniform noise, refers to a noise type that the hidden label flips to a random class with a fixed probability $\eta$, where the true label is included into the label flipping options, which means that in $\eta_{jc}(\mathbf{x}_i)=\frac{\eta}{|\mathcal{Y}|-1}, \forall j \in \mathcal{Y}, \text{ such that } j \neq c$, and $\eta_{cc}(\mathbf{x}_i)=1-\eta$, such that $\eta_{cc} > \frac{1}{|\mathcal{Y}|}$. The theoretical upper bound for the symmetric noise defined by $\eta < \frac{1}{|\mathcal{Y}|}$. 
The asymmetric noise is based on flipping labels between similar classes~\cite{patrini2017making}, where $\eta_{jc}(\mathbf{x}_i)$ depends only on the classes $j,c\in\mathcal{Y}$, but not on $\mathbf{x}_i$, and $\eta_{jc}<\eta_{cc}$ when considering flipping labels between two similar classes.  For example, using  CIFAR-10 data set \cite{krizhevsky2009learning}, the asymmetric noise maps \emph{truck} $\to$ \emph{automobile}, \emph{bird} $\to$ \emph{plane}, \emph{deer} $\to$ \emph{horse}, as mapped by \cite{zhang2018generalized} and $\eta_{jc}<0.5$. 
The instanced-based noise~\cite{rog} depends on both the classes $j,c\in\mathcal{Y}$ and the image $\mathbf{x}_i$.

Below, we first provide details on state-of-the-art (SOTA) noisy-label learning approaches~\cite{li2020dividemix, ding2018semi, kong2019recycling} that we follow, and then we present our two hypotheses to improve the classification accuracy of these approaches.

\subsection{State-of-the-art Noisy-label Learning}

Our algorithm is built upon SOTA noisy-label learning approaches~\cite{li2020dividemix, ding2018semi, kong2019recycling} that are based on: 1) an \textit{unsupervised learning classifier} that characterises training samples as clean or noisy; and 2) an \textit{SSL} classifier that assumes that the training samples classified as clean are labelled, and the samples classified as noisy are unlabelled.
The SOTA noise-robust classifier~\cite{li2020dividemix,nguyen2019self} is formed by an ensemble of two classifiers, each represented by $f:\mathcal{S} \times \Theta \rightarrow [0,1]^{|\mathcal{Y}|}$, where the classifier structure is the same, but their parameters are denoted by $\theta(1),\theta(2) \in \Theta \subset \mathbb{R}^L$.  The training for $\theta(1)$ influences $\theta(2)$ and vice-versa, where this can be achieved by co-training~\cite{li2020dividemix} or student-teacher~\cite{nguyen2019self} approaches. In this paper, we focus on co-training.

\textit{The unsupervised learning classifier} predicts the clean and noisy samples based on their loss values~\cite{arazo2019unsupervised, li2020dividemix,rog, jiang2020beyond}. 
Formally, assuming that the training is minimising the empirical risk 
\begin{equation}
\frac{1}{|\mathcal{D}|}\sum_{i=1}^{|\mathcal{D}|} \ell(f(\mathbf{x}_i;\theta),\mathbf{y}_i),
\label{eq:ERM}
\end{equation}
the set of clean and noisy samples are respectively defined by 
\begin{equation}
\begin{split}
\mathcal{X} &= \left \{ (\mathbf{x}_i,\mathbf{y}_i) : p \left ( \text{clean} | \ell_i , \gamma \right ) \ge \tau \right \}, \\
\mathcal{U} &= \left \{ (\mathbf{x}_i,\mathbf{y}_i^{*}) : 
p \left ( \text{clean} | \ell_i , \gamma \right ) < \tau \right \}, 
\end{split}
\label{eq:set_U_X}
\end{equation}
where $\mathbf{y}_i^{*} = f(\mathbf{x}_i;\theta)$,
$\ell_i = \ell( f (\mathbf{x}_i;\theta),\mathbf{y}_i)$ represents a classification loss (e.g., cross entropy),
and $p \left ( \text{clean} | \ell( f(\mathbf{x}_i;\theta),\mathbf{y}_i) , \gamma \right )$
 is a function that computes the probability that the training sample $(\mathbf{x}_i,\mathbf{y}_i)$ is clean based on its loss $\ell_i$~\cite{jiang2020beyond, li2020dividemix, zhang2020distilling,nguyen2019self}, where this function is parameterised by $\gamma$ (in this paper, this probability function computes the posterior of the smaller-mean component of a bi-modal GMM, where this smaller mean represents the clean GMM component~\cite{li2020dividemix}). To learn $\theta(1)$ and $\theta(2)$, co-training uses the clean and noisy sets from model $\theta(1)$ to train $\theta(2)$, and vice-versa. The classification of samples into clean or noisy, defined in~\eqref{eq:set_U_X}, is done based on the loss value for each training epoch. 
 For learning problems containing large noise rates, this loss value can be unstable over different training epochs, reducing the precision that samples are classified into clean or noisy, resulting in poorer training convergence and generalisation.
 
The \textit{SSL} based on MixMatch~\cite{berthelot2019mixmatch} mixes the elements of $\mathcal{X}$ and $\mathcal{U}$ to minimise the empirical vicinal risk (EVR)~\cite{zhang2017mixup}
\begin{equation}
\small
\begin{split}
\ell_{EVR} =  \frac{1}{|\mathcal{X}'|}\sum_{\substack{(\tilde{\mathbf{x}}_i,\tilde{\mathbf{y}}_i) \\ \in \mathcal{X}'}} \ell^{(\mathcal{X}')}(f(\tilde{\mathbf{x}}_i;\theta),\tilde{\mathbf{y}}_i) + 
 \frac{\lambda^{(\mathcal{U}')}}{|\mathcal{U}'|}\sum_{\substack{(\tilde{\mathbf{x}}_i,\tilde{\mathbf{y}}_i) \\ \in \mathcal{U}'}} \ell^{(\mathcal{U}')}(f(\tilde{\mathbf{x}}_i;\theta),\tilde{\mathbf{y}}_i),
\end{split}
\label{eq:EVR}
\end{equation}
where $\lambda^{(\mathcal{U}')}$ weights the noisy set loss, $\ell^{(\mathcal{X}')}(.)$ and $\ell^{(\mathcal{U}')}(.)$ denote the losses in the clean and noisy sets, defined below in~\eqref{eq:clean_noisy_loss_functions}.  These sets are respectively defined as
\begin{equation}
\small
    \begin{split}
        \mathcal{X}' &= \{ (\tilde{\mathbf{x}}_i,\tilde{\mathbf{y}}_i ) : (\tilde{\mathbf{x}}_i,\tilde{\mathbf{y}}_i ) \sim v(\tilde{\mathbf{x}},\tilde{\mathbf{y}}|\mathbf{x}_i,\mathbf{y}_i), (\mathbf{x}_i,\mathbf{y}_i) \in \mathcal{X}\} \\
        \mathcal{U}' &= \{ (\tilde{\mathbf{x}}_i,\tilde{\mathbf{y}}_i ) : (\tilde{\mathbf{x}}_i,\tilde{\mathbf{y}}_i ) \sim v(\tilde{\mathbf{x}},\tilde{\mathbf{y}}|\mathbf{x}_i,\mathbf{y}_i), (\mathbf{x}_i,\mathbf{y}_i) \in \mathcal{U}\},
    \end{split}
    \label{eq:X_prime_U_prime}
\end{equation}
with
\begin{equation}
\small 
\begin{split}
    v(\tilde{\mathbf{x}},\tilde{\mathbf{y}}|\mathbf{x}_i,\mathbf{y}_i) = \frac{1}{|\mathcal{X} \cup \mathcal{U}|}  \sum_{\substack{(\mathbf{x}_j,\mathbf{y}_j) \\ \in \mathcal{X} \cup \mathcal{U}}} \mathbb{E}_{\lambda} \left [ \delta \left ( \tilde{\mathbf{x}} = \lambda \mathbf{x}_i+(1-\lambda)\mathbf{x}_j,\tilde{\mathbf{y}}=\lambda \mathbf{y}_i + (1-\lambda)\mathbf{y}_j \right ) \right ],
    \end{split}
    \label{eq:v_x_y}
\end{equation}
where $\delta$ is a Dirac mass centered at $(\tilde{\mathbf{x}},\tilde{\mathbf{y}})$, $\lambda \sim \text{Beta}(\alpha,\alpha)$, and $\alpha \in (0,\infty)$.

Note that $v(\tilde{\mathbf{x}},\tilde{\mathbf{y}}|\mathbf{x}_i,\mathbf{y}_i)$ in~\eqref{eq:v_x_y} denotes a distribution that measures the probability of finding the pair $(\tilde{\mathbf{x}},\tilde{\mathbf{y}})$ in the vicinity of $(\mathbf{x}_i,\mathbf{y}_i)$.
The clean set $\mathcal{X}'$ from~\eqref{eq:X_prime_U_prime} is built by sampling the vicinity distribution of clean samples in $\mathcal{X}$, while the noisy set $\mathcal{U}'$ is built by sampling the vicinity distribution of noisy samples from $\mathcal{U}$.
In~\cite{li2020dividemix},  the noisy set size $|\mathcal{U}'|$ and clean set size $|\mathcal{X}'|$ are constrained to be equal to 
$|\mathcal{X}|$, which means that  $|\mathcal{X}'|=|\mathcal{U}'|=|\mathcal{X}|$ , and therefore $|\mathcal{X}'\cup \mathcal{U}'|< |\mathcal{X} \cup \mathcal{U}| $.
The main issue with this constraint is that, for problems with large noise rate, $|\mathcal{X}|$ tends to be small in comparison with the original dataset size $|\mathcal{D}|$, compromising the generalisation of the EVR minimisation in~\eqref{eq:EVR}.

We hypothesise that the classification accuracy of these noisy-label learning methods depends on: 1) the precision of the classification of clean samples to be included in $\mathcal{X}$ in~\eqref{eq:set_U_X} (Section~\ref{sec:hypothesis_one}), and 2) the size of the clean set denoted by $|\mathcal{X}|$ (Section~\ref{sec:hypothesis_two}). 
In particular, a large $|\mathcal{X}|$ with a high proportion of clean samples  will reduce the bound of the difference between the estimated and vicinal risks~\cite{zhang2018generalization}, improving the semi-supervised classification accuracy.

\subsection{Hypothesis One: Improving the Precision in Differentiating Clean and Noisy Samples}
\label{sec:hypothesis_one}

\begin{figure}[h!]
\centering
\includegraphics[width=\textwidth]{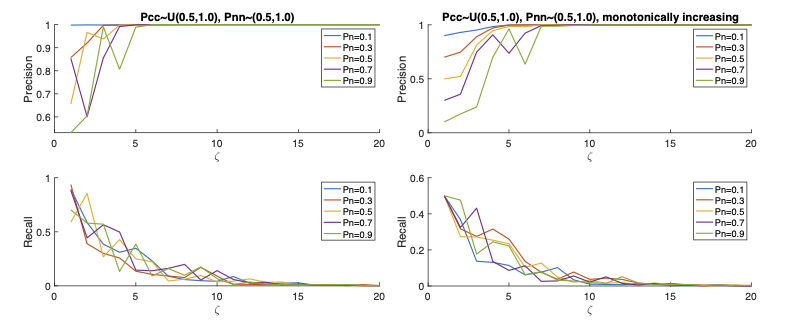}
\caption{Precision and recall as a function of the number of training epochs $\zeta$ (in a range from 1 to 20 in the horizontal axis) for $P^{(e)}_{cc} \sim \mathcal{U}(0.5,1.0)$, $P^{(e)}_{nn} \sim \mathcal{U}(0.5,1.0)$ and $P_n$, where the graph on the left assumes that $P^{(e)}_{cc}$ and $P^{(e)}_{nn}$ are independent between epochs indexed by $e$, while on the right, we assume that $P^{(e_1)}_{cc} < P^{(e_2)}_{cc}$ and $P^{(e_1)}_{nn} < P^{(e_2)}_{nn}$ for epoch $e_2 > e_1$.}
\label{fig:motivation_proof}
\end{figure}

We first hypothesise that the precision in the classification of clean samples in $\mathcal{X}$ can be improved by classifying as clean the samples that consistently show $p(\text{clean}|\ell_{i},\gamma) \geq \tau$ for $\zeta$ epochs.  
We conjecture that such improved precision leads to a higher classification accuracy.

Assuming that $P^{(e)}_{cc}$ denotes the probability of classifying a clean sample as clean during training epoch $e \in \{1,...,E\}$, $P^{(e)}_{nc} = 1 - P^{(e)}_{cc}$ the probability of classifying a clean sample as noisy. Similarly, $P^{(e)}_{nn}$ represents the probability of classifying a noisy sample as noisy during training epoch $e$, $P^{(e)}_{cn}=1-P^{(e)}_{nn}$ the probability of classifying a noisy sample as clean. Also, $P_c$ and $P_n$ denote the proportion of clean and noisy samples in the training set, with $P_n+P_c=1$.  The probability of a clean sample being in the clean set $\mathcal{X}$ for $\zeta$ epochs is $\prod_{e=t+1}^{t+\zeta} P^{(e)}_{cc}$, and 
the probability of a noisy sample being in the clean set for $\zeta$ epochs is $\prod_{e=t+1}^{t+\zeta} P^{(e)}_{cn}=\prod_{e=t+1}^{t+\zeta}(1-P^{(e)}_{nn})$.

\begin{lemma}
\label{lemma:zeta}

Assuming that $P^{(e)}_{cc} \sim \mathcal{U}(0.5,1.0)$ (so $P^{(e)}_{nc} = (1-P^{(e)}_{cc}) \in (0.0,0.5)$) and $P^{(e)}_{nn} \sim \mathcal{U}(0.5,1.0)$ (so $P^{(e)}_{cn} = (1-P^{(e)}_{nn}) \in (0.0,0.5)$), the classification precision of clean samples in $\mathcal{X}$ tends to 1 and recall tends to 0, as $\zeta$ increases, with $\mathcal{U}(a,b)$ denoting the uniform distribution between $a$ and $b$.

\end{lemma}

\begin{proof}

The precision and recall are calculated with:
\begin{equation}
\begin{split}
    \text{Precision}&=\frac{P_c \times  \prod_{e=t+1}^{t+\zeta} P^{(e)}_{cc}}{ P_c \times \prod_{e=t+1}^{t+\zeta} P^{(e)}_{cc} +  P_n \times \prod_{e=t+1}^{t+\zeta} P^{(e)}_{cn}}, \\
    \text{Recall}&=\frac{ P_c \times \prod_{e=t+1}^{t+\zeta} P^{(e)}_{cc}}{ P_c \times \prod_{e=t+1}^{t+\zeta} P^{(e)}_{cc} + P_c \times  (1-\prod_{e=t+1}^{t+\zeta}P^{(e)}_{cc})},
\end{split}
\label{eq:precision_recall}
\end{equation}
where $\text{Precision} = \frac{TP}{TP+FP}$ and $\text{Recall}=\frac{TP}{TP+FN}$, with true positives ($TP$) being computed by the proportion of clean samples $P_c$ multiplied by the proportion of clean samples classified as clean $P_{cc}^{(e)}$, false positives ($FP$) being calculated with the proportion of noisy samples $P_n$ times the proportion of noisy samples classified as clean $P_{cn}^{(e)}$, and false negatives ($FN$) being computed as the proportion of clean samples $P_c$ times the proportion of clean samples classified as noisy $P_{nc}^{(e)}$ -- all terms are computed over $\zeta$ epochs.
Given that $P^{(e)}_{cn} \in (0.0,0.5)$ and $P^{(e)}_{cc} \in (0.5,1.0)$ and that
\noindent$\lim_{\zeta \rightarrow \infty} (\prod_{e=t+1}^{t+\zeta} P^{(e)}_{cn}/\prod_{e=t+1}^{t+\zeta} P^{(e)}_{cc})\rightarrow 0$,  $\text{Precision}$ tends to 1, and similarly, given that $\lim_{\zeta \rightarrow \infty} (1-\prod_{e=t+1}^{t+\zeta}P^{(e)}_{cc})/\prod_{e=t+1}^{t+\zeta} P^{(e)}_{cc})\rightarrow \infty$,  $\text{Recall}$ tends to 0.

\end{proof}

Figure~\ref{fig:motivation_proof} shows precision and recall 
as a function of the number of training epochs $\zeta$ (in a range from 1 to 20 in the horizontal axis)
for several values of $P_n$ and assuming that $P^{(e)}_{cc}$ and $P^{(e)}_{nn}$ are independent between epochs (left), while on the right, we assume that $P^{(e_1)}_{cc} < P^{(e_2)}_{cc}$ and $P^{(e_1)}_{nn} < P^{(e_2)}_{nn}$ for $e_2 > e_1$. 
As demonstrated in Lemma~\ref{lemma:zeta}, precision increases to 1 and recall decreases to 0 as we increase $\zeta$.

 Section~\ref{sec:hypothesis_one_empirical_results} shows empirical evidence that our approach to improve the precision when differentiating between clean and noisy-label samples, increases the clean set precision and classification accuracy, compared with the approach based on the small-loss strategy~\cite{li2020dividemix}. Although the addition of noisy samples in the clean set might help the performance in some noisy-label scenarios, it is hard to measure the amount and to select the noisy samples that can increase classification accuracy. In general, in scenarios where the noisy-label rate is high, the amount of noisy-label samples that are predicted as clean and inserted into the labelled data can also be high, which has a negative impact on classification accuracy, as shown in Section~\ref{sec:hypothesis_one_empirical_results}. 
 Therefore, targeting the formation of a clean set that has the highest possible proportion of clean samples constitutes our main goal when building the clean set, as explained in this section.

\subsection{Hypothesis Two: Oversampling the Clean Data to Increase the Robustness of EVR}
\label{sec:hypothesis_two}

We also hypothesise that the increase of the clean set $|\mathcal{X}|$ leads to a decrease of the bound for vicinal risk minimisation, which improves the semi-supervised classification accuracy, as shown in Theorem 8 of the paper~\cite{zhang2018generalization}.
More specifically, for low noise rate problems, $P_{cc}$ tends to be large and $P_{cn}$, small, so even for small values of $\zeta$, precision will be close to one with a relatively high Recall, allowing for a large $|\mathcal{X}|$ (see blue curves in Fig.~\ref{fig:motivation_proof} for $P_n=0.1$).
On the other hand, $P_{cc}$ tends to be small and $P_{cn}$ large in high noise rate scenarios, which means that $\zeta$ needs to increase to push the precision to be close to one, but that can reduce the Recall to very low values, resulting in a potentially small $|\mathcal{X}|$.  
Therefore, $\zeta$ is a hyper-parameter that needs to be estimated to enable high precision and large $|\mathcal{X}|$.
Nevertheless, even with a careful estimation of $\zeta$, pushing the precision to be high in large noise rate scenarios can still result in a small $|\mathcal{X}|$.

Hence, we propose two solutions to address the small $|\mathcal{X}|$ issue. The first solution consists of dividing the training process into two stages, where the first stage finds the largest possible set of high-confidence clean samples $\mathcal{X}$, and the second stage trains the model using $\mathcal{X}$ as the minimum set of clean samples that is augmented with clean samples from the small-loss strategy. The second solution consists of sampling $\mathcal{X}$
 with replacement when mixing up $\mathcal{X}'$ and $\mathcal{U}'$ in~\eqref{eq:X_prime_U_prime}, such that $|\mathcal{X}'|=|\mathcal{U}'|=|\mathcal{D}|$, which decreases the vicinal risk minimisation bound~\cite{zhang2018generalization}, improving the semi-supervised classification accuracy. Differently from standard oversampling approaches, which focus on increasing the minority class $\mathcal{X}'$ (for high noise scenarios), our focus is to increase the Mixup iterations between $\mathcal{X}'$ and $\mathcal{U}'$ rather than just increasing the size of $\mathcal{X}'$.

 In Section~\ref{sec:hypothesis_two_empirical_results}, we present an empirical results in Figure~\ref{fig:acc_iters} that shows that the test accuracy, as a function of training steps (iterations), for LongMix is better than DivideMix~\cite{li2020dividemix}. In other words, this shows that adding more MixUp iterations per epoch, as in LongMix, is not equivalent to adding more epochs, as in the DivideMix baseline~\cite{li2020dividemix}, so an  increase in the number of epochs is not equivalent to adding more MixUp iterations, as we propose for LongMix. 
 Furthermore, by fixing the number of training iterations for LongReMix and DivideMix, we show in Table~\ref{tab:longmix_sameiter} that our approach produces higher test accuracy, particularly for the high-noise rate scenarios. We also show in Table~\ref{tab:longmix_sameiter} that LongMix achieves higher accuracy results than the baseline using oversampling of $\mathcal{X}'$.

\section{LongReMix}
\label{sec:LongReMix}

In this section, we present the main contribution of this paper, namely the LongReMix noisy-label learning algorithm.
The two hypothesis in 
Sections~\ref{sec:hypothesis_one} and~\ref{sec:hypothesis_two}
are explored for developing LongReMix.
A simplified diagram of LongReMix is displayed in 
Figures~\ref{fig:motivation_part1} and~\ref{fig:motivation_part2},
where the first stage comprises the High Confidence Training (HCT), which trains the model to find a high confidence set of clean samples with high precision. 
In the second stage, we combine this high confidence set of clean samples with the clean samples from the small-loss strategy to retrain the model. 
This retraining uses a new way to build the data sets $\mathcal{X}'$ and $\mathcal{U}'$ in~\eqref{eq:X_prime_U_prime}, called LongMix,
which enables the number of MixUp operations to be proportional to $|\mathcal{D}|$ instead of  $|\mathcal{X}|$, as described in 
Section~\ref{sec:hypothesis_two}. Therefore, the number of additional MixUp operations will be equal to $|\mathcal{D}|-|\mathcal{X}|$. 
LongMix is the training strategy which  oversamples the clean set, whereas LongReMix is the two-stage training, which uses HCT in the first stage and LongMix in the second stage.

\subsection{First Stage: High Confidence Training}
\label{sec:HCT}

The \textbf{high confidence training (HCT)} stage aims to increase the precision, without reducing too much the recall, of the unsupervised classification of clean and noisy training samples.  
Following the ideas presented in Sections~\ref{sec:hypothesis_one} and~\ref{sec:hypothesis_two},
we re-define how to form the sets of clean and noisy samples, originally defined in~\eqref{eq:set_U_X}, as follows:
\begin{equation}
\small
    \begin{split}
\mathcal{X}_1^{(e)} &= \left \{ (\mathbf{x}_i,\mathbf{y}_i,w_i) : 
w_i=p  ( \text{clean} |   \ell_i^{(e)}  , \gamma  ) \ge \tau, \forall e \in \mathcal{E} \right \}, \\
\mathcal{U}_{1}^{(e)} &= \left \{ (\mathbf{x}_i,\mathbf{y}_i^{*},w_i) : 
w_i=p  ( \text{clean} | \ell_i^{(e)} , \gamma ) < \tau , \exists e \in \mathcal{E} \right \}, 
    \end{split}
\label{eq:redefine_clean_noisy_sets}
\end{equation}
where $\ell_i^{(e)}$ represents the loss of sample $(\mathbf{x}_i,\mathbf{y}_i)$ at training epoch $e$ and $\mathcal{E}$ denotes the confidence window comprising the current and the previous $(\zeta-1)$ epochs -- this is represented by the block ``filter''  that produces the high 
confidence clean set in Figure~\ref{fig:motivation_part1}.
Hence, \emph{a sample to be in the clean set $\mathcal{X}_1^{(e)}$ must be classified as clean for $\zeta$ epochs in a row}, resulting in a more consistent, but smaller, set of clean samples, containing fewer noisy samples than the set in~\eqref{eq:set_U_X}.

\subsection{Second Stage: Guided Training}
\label{sec:guided_training}

The \textbf{Guided Training} stage depends on the high-confidence set of clean samples estimated from the first training stage with
\begin{equation}
    \mathcal{H} = \arg \max_{\mathcal{X}_{1}^{(e)}:e \in \{ \frac{E}{2},...,E\} }  |\mathcal{X}_{1}^{(e)}|,
    \label{eq:H}
\end{equation}
where $E$ is the total number of training epochs for the first stage of training. 
Therefore, the high confidence set $\mathcal{H}$ from~\eqref{eq:H} consists of the largest clean set $\mathcal{X}_1^{(e)}$ obtained from the second half (i.e., $e \in \{ \frac{E}{2},...,E\}$) of the first stage of training (i.e., the HCT training stage).
In the second stage of training, we define the labelled and unlabelled sets as in~\eqref{eq:set_U_X}, but we use $\mathcal{H}$ to update these sets as follows:
\begin{equation}
\scalebox{0.85}{
$
\begin{split}
\mathcal{X}_2^{(e)} =&
\Big \{  (\mathbf{x}_i,\mathbf{y}_i,w_i) : 
\left\{\begin{array}{l}
w_i=1 ,\textrm{if} \left (\mathbf{x}_i,\mathbf{y}_i \right) \in \mathcal{H}; \text{ or} \\ 
w_i=p \left( \textrm{clean} |   \ell_i^{(e)}  , \gamma \right)\ge \tau ,\textrm{otherwise} 
\end{array}\right.\Big \},\\
 \mathcal{U}_2^{(e)} = & \Big \{  (\mathbf{x}_i,\mathbf{y}_i^{*},w_i) : 
w_i=p  \left( \text{clean} | \ell_i^{(e)} , \gamma \right ) < \tau,   \text{ and }  (\mathbf{x}_i,\mathbf{y}_i) \notin \mathcal{H}   \Big \}.
\end{split}
$
}
\label{eq:redefine_2_clean_noisy_sets}
\end{equation}
Hence, to form the clean set $\mathcal{X}_2^{(e)}$ in the second training stage, samples need to be in the high-confidence set $\mathcal{H}$ from~\eqref{eq:H} or be classified as clean using the small loss strategy (i.e., $p( \textrm{clean} |   \ell_i^{(e)}  , \gamma )\ge \tau$).  To form the noisy set $\mathcal{U}_2^{(e)}$, samples cannot be in the high-confidence set $\mathcal{H}$ and they have to be classified as noisy with $p ( \textrm{clean} |   \ell_i^{(e)}  , \gamma ) < \tau$.
During the second stage of LongReMix, we retrain the model from scratch\footnote{We compared 
if we should fine-tune the model trained from the first stage or train from scratch, and the latter approach showed the best results.} using the
clean and noisy samples defined in~\eqref{eq:redefine_2_clean_noisy_sets}.


As explained in 
Sections~\ref{sec:hypothesis_one} and~\ref{sec:hypothesis_two},
we hypothesise that by sampling the clean set with replacement (i.e. oversampling the clean set), we increase the number of MixUp operations in the EVR loss in~\eqref{eq:EVR}, resulting in a smaller bound of the difference between estimated and vicinal risks~\cite{zhang2018generalization}.  
Therefore, we propose \textbf{LongMix} that increases the number of MixUp operations to be $|\mathcal{D}|$, instead of the number of predicted clean samples. 
A criticism faced by LongMix is that adding more MixUp iterations per epoch may be equivalent to a simple increase in the number of epochs, but we show in the experiments that this is not true.

\subsection{Training and Inference}

The training loss for our proposed LongReMix is~\cite{li2020dividemix}:
\begin{equation}
    \ell = \ell_{EVR} + \lambda_{R}\ell_{R},
    \label{eq:D_loss_full}
\end{equation}
where $\ell_{EVR}$ denotes the empirical vicinal error from~\eqref{eq:EVR}, which sums the loss functions from the clean set, represented by $\ell^{(\mathcal{X}')}$, and from the noisy set, denoted by $\ell^{(\mathcal{U}')}$, which are defined as
\begin{equation}
    \begin{split}
        \ell^{(\mathcal{X}')}(f(\tilde{\mathbf{x}}_i;\theta),\tilde{\mathbf{y}_i}) & =-\tilde{\mathbf{y}}_i^{\top}\log(f(\tilde{\mathbf{x}}_i;\theta)), \\ \ell^{(\mathcal{U}')}(f(\tilde{\mathbf{x}}_i;\theta),\tilde{\mathbf{y}_i}) & = \| \tilde{\mathbf{y}}_i - f(\tilde{\mathbf{x}}_i;\theta) \|^2_2,
    \end{split}
    \label{eq:clean_noisy_loss_functions}
\end{equation}
$\lambda_{R}$ weights the regularisation loss defined as
\begin{equation}
\small
    \ell_{R} =
    KL \left [ \pi_{|\mathcal{Y}|} \Bigg | \Bigg | \frac{1}{|\mathcal{X}'| + |\mathcal{U}'|} \sum_{\mathbf{x} \in (\mathcal{X}' \bigcup \mathcal{U}')} f(\mathbf{x};\theta) \right ],
    \label{eq:L_reg}
\end{equation}
which regularizes the training by approximating the model output for all samples to a uniform distribution, with $\pi_{|\mathcal{Y}|}$ denoting a vector of $|\mathcal{Y}|$ dimensions with values equal to $1/|\mathcal{Y}|$, and 
$KL[a||b]$ representing the Kullback Leibler divergence between $a$ and $b$.

The pseudocode for the training of LongReMix is shown in Algorithm~\ref{alg:LRM}, where the function $\text{DataAugment}(\mathbf{x},M)$ returns $M$ augmented samples from $\mathbf{x}$ using simple geometric transformations; and 
$\text{TempShrp}(\mathbf{y},T)$ sharpens the distribution $\mathbf{y}$ with temperature $T$.
The inference for a test sample is calculated based on the average of predictions from both networks. Our model has the same number of hyper-parameters and the same values as SOTA methods~\cite{li2020dividemix}, except for the addition of the confidence window $\zeta$. All hyper-parameters from Algorithm~\ref{alg:LRM} are fixed independently of the data set.



\begin{algorithm}[t!]
\scriptsize

\KwInput{$\mathcal{D}$, batch size $B$, threshold $\tau$, number of augmentations $M$, temperature sharpening $T$, $\lambda^{(\mathcal{U})}$, $\lambda_{R}$, Beta parameter $\alpha$, confidence window $\zeta$,  number of epochs for warm-up ($E_{W}$) and training ($E$),  $stage \in \{1,2\}$, and $\mathcal{H}$ (if $stage==2$).}
Pre-train $f(\mathbf{x};\theta(1))$ and $f(\mathbf{x};\theta(2))$ with cross entropy (CE) loss in $\mathcal{D}$ for $E_{W}$ iterations.\\
 \While{$e < E$}{
 \For{$i = \{1,...,|\mathcal{D}|\}$}
 {Estimate $p(\text{clean}|\ell^{(e)}_i(1),\gamma)$,
 $p(\text{clean}|\ell^{(e)}_i(2),\gamma)$, where $\ell^{(e)}_i(k)$ is the CE loss for $(f(\mathbf{x}_i;\theta(k)),\mathbf{y}_i)$}
  \For{k=1,2}   
  {
    \eIf{$stage == 1$}{
        Build sets $\mathcal{X}_{stage}^{(e)}(k)$ and $\mathcal{U}_{stage}^{(e)}(k)$ using~\eqref{eq:redefine_clean_noisy_sets} with  $p(\text{clean}|\ell^{(e)}_i(\text{mod}(k,2)+1),\gamma)$\\
        $num\_iters$ = $\frac{|\mathcal{X}_{stage}^{(e)}(k)|}{B}$
    }
    {
    Build sets $\mathcal{X}_{stage}^{(e)}(k)$ and $\mathcal{U}_{stage}^{(e)}(k)$ using~\eqref{eq:redefine_2_clean_noisy_sets} with  $p(\text{clean}|\ell^{(e)}_i(\text{mod}(k,2)+1),\gamma)$ and $\mathcal{H}$\\
        $num\_iters$ = $\frac{|\mathcal{D}|}{B}$
    }
    \For{iter=1 \textbf{to} $num\_iters$}   
    {
        $\{(\mathbf{x}_b, \mathbf{y}_b, w_b)\}_{b=1}^{B} \sim \mathcal{X}_{stage}^{(e)}(k)$\\
        $\{(\mathbf{u}_b, \mathbf{y}^*_b, w_b)\}_{b=1}^{B} \sim \mathcal{U}_{stage}^{(e)}(k)$
        
        \For{b=1 \textbf{to} $B$}   
        {
            $\{ \hat{\mathbf{x}}_{b,m}\}_{m=1}^{M}$ = DataAugment($\mathbf{x}_b,M$)\\
            $\{ \hat{\mathbf{u}}_{b,m} \}_{m=1}^{M}$ = DataAugment($\mathbf{u}_b,M$)
                
            $\mathbf{p}_b =  \frac{1}{M}\sum\limits_{m=1}^{M}
            f(\hat{\mathbf{x}}_{b,m};\theta(k))$\\
            $\mathbf{q}_b = \frac{1}{2M}\sum\limits_{m=1,l=1}^{M,2} f(\hat{\mathbf{u}}_{b,m};\theta(l)))$
            
            $\hat{\mathbf{y}}_b = \text{TempShrp}(w_b \mathbf{y}_b + (1-w_b)\mathbf{p}_b;T)$\\
            $\hat{\mathbf{q}}_b = \text{TempShrp}(\mathbf{q}_b;T)$
            
        }
        Build sets $\mathcal{X}'$,$\mathcal{U}'$
        using~\eqref{eq:X_prime_U_prime} with $\mathcal{X}'=\{(\hat{\mathbf{x}}_{b,m},\hat{\mathbf{y}}_b)\}_{b \in (1,...,B), m \in (1,...,M)}$ 
        $\mathcal{U}'=\{(\hat{\mathbf{u}}_{b,m},\hat{\mathbf{q}}_b) \}_{b \in (1,...,B), m \in (1,...,M)}$
        Update $\theta(k)$ with $\ell$ from~\eqref{eq:D_loss_full}
        
    }
    }
 }
 \KwOutput{$\theta(1), \theta(2)$}
 \caption{LongReMix}
 \label{alg:LRM}
 
\end{algorithm}

Assuming that we have 2 models for training, $E$ is the number of epochs, $M$ denotes the number of data augmentations, $|\mathcal{D}|$ represents the training set size, and $L$ is the number of model parameters, 
the run-time complexity of LongReMix is dominated by the main training iteration of the first (high confidence training) and second (guided training) stages, which is as follows: $\mathcal{O}(2 \times 2 \times E \times M \times |\mathcal{D}| \times L)$. The original time complexity of noisy-label SSL methods with two models is $\mathcal{O}(2 \times E \times M \times |\mathcal{D}| \times L)$. Hence, LongReMix has a training process that is roughly two times longer than the original noisy-label SSL methods, but both LongReMix and noisy-label SSL methods are asymptotically linear in all parameters.

\section{Experiments}\label{sec:experiments}

We compare LongReMix with related approaches on five noisy-label learning benchmarks. We also analyze the performance of LongReMix on a number of ablation studies, where we show empirical evidence of the hypotheses in Sections~\ref{sec:hypothesis_one} and~\ref{sec:hypothesis_two}. 

\subsection{Data Sets}
\label{sec:datasets}

We conduct our experiments on the data sets  CIFAR-10, CIFAR-100~\cite{krizhevsky2009learning}, CNWL (Red Mini-ImageNet)~\cite{jiang2020beyond},
Clothing1M~\cite{xiao2015learning}, WebVision~\cite{li2017WebVision} and Food101-N~\cite{lee2018cleannet}. These datasets follow the main evaluation protocol for noisy labels used in literature~\cite{li2020dividemix,ren2018learning,shu2019meta}, which covers symmetric, asymmetric and real-world instance-dependent noise. CIFAR-10 and CIFAR-100 have 50000 training and 10000 testing images of size $32 \times 32$ pixels, where CIFAR-10 has 10 classes and CIFAR-100 has 100 classes and all training and testing sets have a perfectly balanced number of images per classes. As CIFAR-10 and CIFAR-100 data sets originally do not contain label noise, a common approach is to add synthetic noise to evaluate the models. For CIFAR-10/CIFAR-100 we investigated three  noise types: symmetric, asymmetric and instanced-based, as defined in Section~\ref{sec:problem_definition}. 
The symmetric noise is generated using $\eta \in \{0.2, 0.5, 0.8, 0.9\}$, with $\eta$  defined in Section~\ref{sec:problem_definition}.

 The asymmetric noise for CIFAR-10 is produced following the mapping used in~\cite{li2020dividemix}, which maps the classes \emph{truck} $\to$ \emph{automobile}, \emph{bird} $\to$ \emph{plane}, \emph{deer} $\to$ \emph{horse}, with $\eta_{jc} \in \{0.4, 0.49\}$ (note that we study $\eta_{jc}=49\%$ because it is close to the theoretical limit of 50\% for this type of noise). The asymmetric noise for CIFAR-100 is produced following the mapping used in~\cite{patrini2017making}, which groups the 100 classes into 20 super-classes containing 5 original classes (e.g., super-class 'Aquatic Mammals' contains 'Beaver', Dolphin', 'Otter', 'Seal', and 'Whale'), and within each super-class the noise flips each class into the next one, circularly.  We also evaluate the instanced-based noise scenario, where 
 we follow the setup from~\cite{rog} to generate semantic (or instance-dependent) noisy labels using a trained VGG~\cite{vgg}, DenseNet~(DN), and ResNet~(RN) on CIFAR-10 and CIFAR-100. This instance-dependent noise was generated by training the VGG, DN and RN using 5\% of the clean-labelled samples from CIFAR-10 and 20\% from CIFAR-100 (also clean-labelled samples), and producing the labels of the remaining training samples from the trained model predictions. 
 The instance-dependent label noise dataset for CIFAR-10 and CIFAR-100 are separated by the predictions of each model.

The CNWL dataset~\cite{jiang2020beyond} forms a benchmark in the study of real-world web label noise. The dataset is built with images and labels being crawled from the web, where the noisy label samples are represented by the matching images. 
The amount of noise is controlled, with amounts varying from 0\% to 80\%, where we specifically use noise rates 20\%, 60\% and 80\%, as suggested by~\cite{FaMUS}. 
The Red Mini-ImageNet comprises 100 classes with 50000 training images and 5000 test images, where the original 84$\times$84-pixel images are resized to 32$\times$32 pixels.

Clothing1M consists of 1 million training images acquired from online shopping websites and it is composed of 14 classes.
As the images from the data set vary in size, we resized the images to $256 \times 256$ for training, as used in \cite{li2020dividemix, han2019deep}.
The data set is heavily imbalanced and most of the noise is asymmetric~\cite{yi2019probabilistic}, with noise rate estimated to be around 40\%~\cite{xiao2015learning}. The data set provide additional clean sets for 
training, validation, and test of 50k, 14k and 10k images, respectively. For our experiments we do not use any of the clean training or validation sets, but we use the test set for evaluation.

WebVision contains 2.4 million images collected from the internet, with the same 1000 classes from ILSVRC12~\cite{deng2009imagenet} and images resized  to $256 \times 256$ pixels. It provides a clean test set of 50k images, with 50 images per class. We compare our model using the first 50 classes of the Google image subset, as used in \cite{li2020dividemix, chen2019understanding}.

Food101-N~\cite{lee2018cleannet} contains 310,009 training images of food recipes classified in 101 classes and 25,000 images for the testing set. The images from this data set were resized to $256 \times 256$. 
This data set is based on the Food101 data set~\cite{bossard2014food}, but it has more images with noisy labels. 
The test set is the same provided by the original Food101~\cite{bossard2014food}, which is a clean test set of 25K images.

\subsection{Implementation}

The model $f(\mathbf{x};\theta)$ is represented by a 18-layer PreAct ResNet18 (PRN18)~\cite{he2016identity} 
 for CIFAR-10 and CIFAR-100, InceptionV2~\cite{szegedy2017inception} for WebVision (this is the model used by competing approaches), and ResNet50~\cite{he2016deep} for Clothing1M and Food-101N. The  PreAct ResNet18 (PRN18) used for CIFAR-10 and CIFAR-100, is a 18-layer Residual Network (ResNet) containing convolutional and pooling layers, and residual blocks, where skip connections are added to deal with the vanishing gradient issue. It also has a pre-activation variant of residual block, where the ReLU layer is moved from the shortcut connection path to an earlier layer. The ResNet50, used for Clothing-1M and Food 101N, is a 50-layer Residual Network,  with standard residual blocks without pre-activation. InceptionV2 is a CNN composed of inception modules, which consist of wider layers to compute convolutions of different filter sizes that are concatenated and sent to the next inception module. The models for each data set were selected according to the evaluation protocol used in prior works \cite{li2020dividemix, li2019learning, arazo2019unsupervised, yu2019does, xiao2015learning}. 
The models are trained with stochastic gradient descent with momentum of 0.8, weight decay of 0.0005 and batch size of 64. The learning rate is 0.02 which is reduced to 0.002 in the middle of the training. 

The WarmUp and total number of epochs is defined according to each data set, as defined in~\cite{li2020dividemix}.
For CIFAR-10 and CIFAR-100, PRN18 is based on a WarmUp stage of 30 epochs, with 300 epochs of total training.
For WebVision, the InceptionV2 is trained for 100 epochs, with a WarmUp stage of 1 epoch.
For Clothing1M, ResNet-50 is trained for 80 epochs with WarmUp stage of 1 epoch.
For Food-101N, we also use ResNet-50 and rely on the same training protocol as in~\cite{han2019deep}, consisting of training for 30 epochs, WarmUp stage of 1 epoch and reducing the learning rate by a factor of 10 every 10 epochs.

The Mixup parameter is $\alpha=4$, used to estimate $\lambda \sim \text{Beta}(\alpha,\alpha)$ in~\eqref{eq:v_x_y}, and 
the regularisation weight for the loss in~\eqref{eq:D_loss_full} is $\lambda_{R}=1$ for symmetric noise and $\lambda_{R}=0$ for asymmetric noise--these two parameters are as defined in~\cite{li2020dividemix}.

\begin{table}[ht]
\centering
\scriptsize
\begin{tabular}{cl||cc|cc}
\toprule
\multicolumn{2}{c}{noise type$\rightarrow$}& \multicolumn{2}{c}{sym.}& Asym. \\
\midrule
\multicolumn{2}{c}{$\zeta\downarrow$, noise rate$\rightarrow$} & 80\% & 90\% & 40\% & 49\% \\
\midrule
\multirow{4}{*}{1}& Precision  &94.06 & 58.61 & 97.38& 79.84 \\ 
& Recall  & \textbf{95.88} &  \textbf{92.18}& \textbf{94.24}& \textbf{96.95} \\ 
& Ac. (Best) & \textbf{94.14} &80.07 & 94.43& 81.94\\ 
& Ac. (Last)  &93.62 & 79.11 & 94.03& 72.73\\ 
\midrule
\multirow{4}{*}{3}& Precision &\textbf{97.53} &70.12 &98.92 & 87.24\\ 
& Recall  &93.14 &80.97 & 89.27 & 93.49\\ 
& Ac. (Best) &94.08 &81.63 & 94.75 & 84.01\\ 
& Ac. (Last) &\textbf{93.70} &80.92 & \textbf{94.44}&78.36\\ 
\midrule
\multirow{4}{*}{5}& Precision &97.39 &\textbf{75.47} & 99.11& 88.05\\ 
& Recall  &89.79 & 84.59& 87.45& 91.86\\ 
& Ac. (Best)  &93.87 &\textbf{83.38} & \textbf{94.83}& 85.12\\ 
& Ac. (Last)  &93.36 &\textbf{82.81} & 94.33& 81.01\\ 
\midrule
\multirow{4}{*}{10}& Precision & \textbf{97.50} &72.40 & \textbf{99.35}& \textbf{88.91}\\ 
& Recall & 84.75 &78.08 & 82.94&89.18\\ 
& Ac. (Best)  &93.94 &82.28 & 94.61& \textbf{85.94}\\ 
& Ac. (Last) &93.45 &81.42 & 94.11& \textbf{82.91}\\ 
\bottomrule
\end{tabular}
\caption{Results of Precision, Recall and Accuracy (Best value and avg. of Last 10 epochs) of LongReMix for different values of $\zeta$, for CIFAR-10 with symmetric and asymmetric noise. We highlight the highest precision, recall, and accuracy (best and last) values for each noisy label problem.}
\label{tab:confidence_window}
\end{table}

\begin{figure}[t]
\centering
\includegraphics[width=0.5\columnwidth]{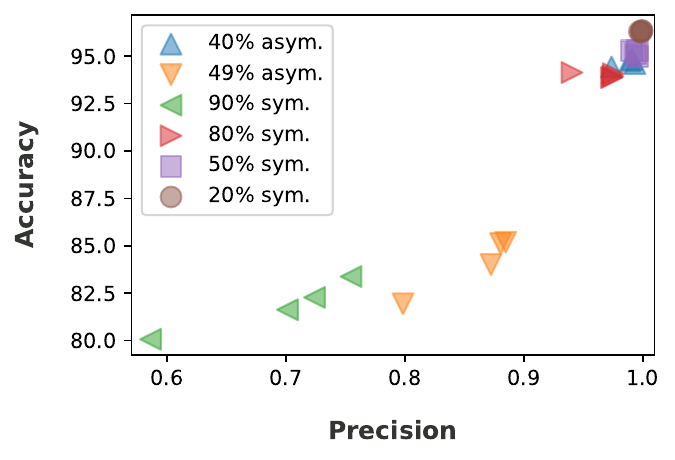}
\caption{Best test accuracy versus the precision of predicted clean set for CIFAR-10 for different levels and types of label noise, using LongReMix. The different values of precision are obtained from  different values of $\zeta\in\{1,3,5,10\}$. All the runs were performed using 300 epochs. 
}
\label{fig:prec_acc}
\end{figure}

\begin{figure}[t]
\centering
\subfloat[Training Loss vs. Epochs]{\includegraphics[width=0.4\columnwidth]{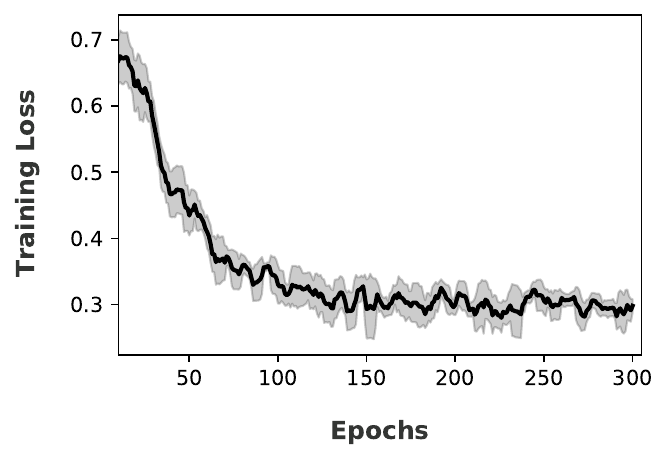} }%
\quad\quad
\subfloat[Precision of predicted clean set vs. Epochs]{
\includegraphics[width=0.4\columnwidth]{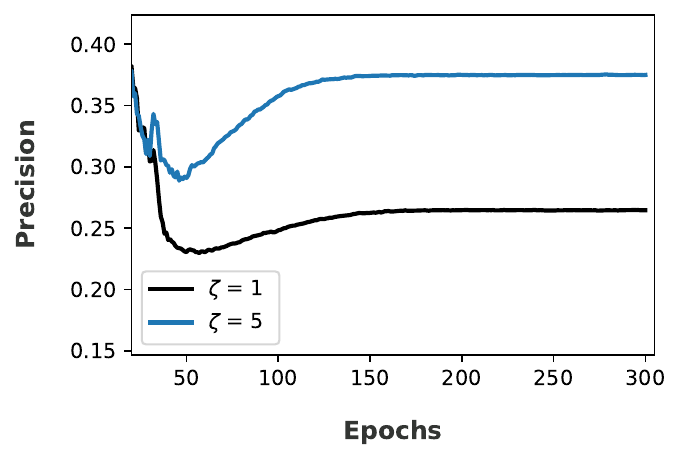}
}
\caption{Training loss versus number of epochs for CIFAR-100 at 90\% symmetric noise rate, using HCT, where the black line and gray error bar denote the mean and standard deviation of the training loss of the last 10 epochs (a), and the precision of the clean set as the first stage training progresses, using $\zeta \in \{1,5\}$ (b).
}
\label{fig:loss_epoch}
\end{figure}

\begin{figure}[ht!]
\centering
\subfloat[Precision vs. Recall for $40\%$ asymmetric noise.]{\includegraphics[width=0.4\columnwidth]{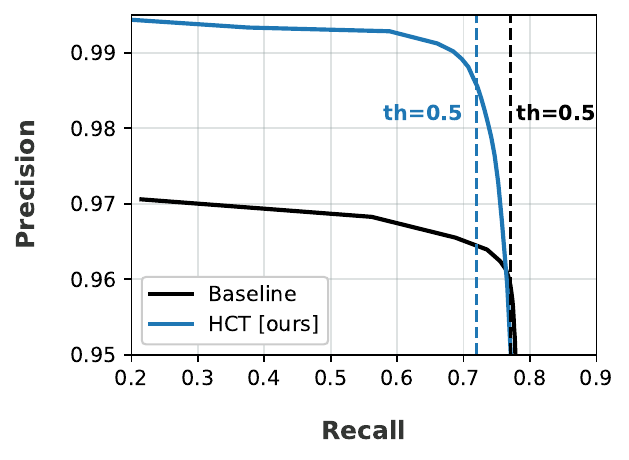} }%
\quad\quad
\subfloat[Precision vs. Noise rate]{
\includegraphics[width=0.5\columnwidth]{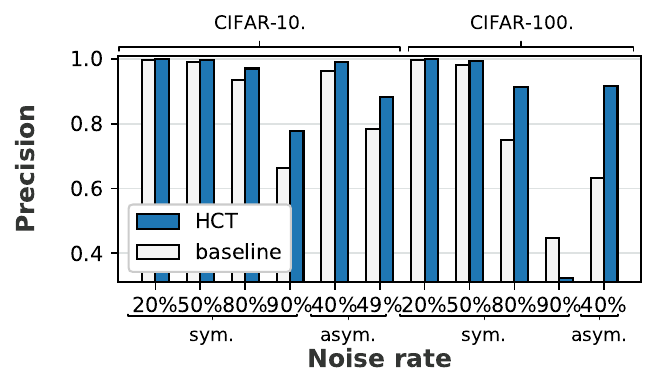}
}
\quad
\subfloat[Recall vs. Noise rate]{
\includegraphics[width=0.5\columnwidth]{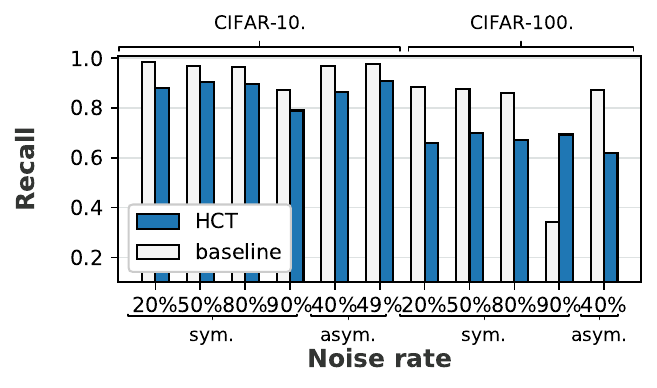} 
}
\subfloat[Number of clean samples vs. Noise rate]{
\includegraphics[width=0.5\columnwidth]{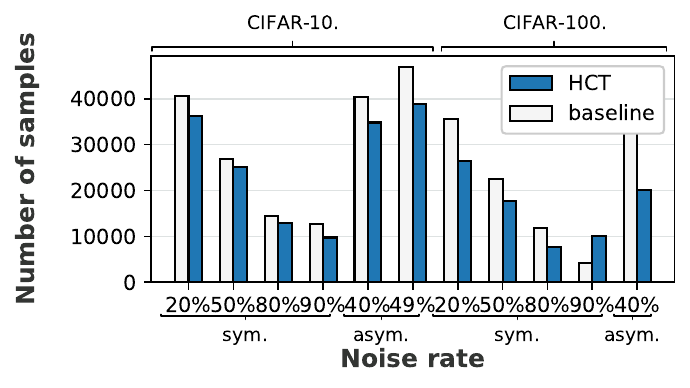}
}

\caption{(a) Precision versus Recall for our proposed $\mathcal{X}_1^{(e)}$ from~\eqref{eq:redefine_clean_noisy_sets} (denoted by HCT) and $\mathcal{X}$ from~\eqref{eq:set_U_X} (Baseline~\cite{li2020dividemix}), for 40\% asymmetric noise on CIFAR-10, where $\tau \in [0,1]$ (denoted by \emph{th}) for $p(\text{clean}|\ell,\gamma)$. (b) Precision  and (c) Recall and (d) number of clean samples for different noise rates, for CIFAR-10 and CIFAR-100, using $\tau =0.5$.}
\label{fig:precison}
\end{figure}

\begin{table}[ht]
\centering
\footnotesize
\scalebox{0.8}{
\begin{tabular}{cc|cccc|cc||cccc}
\toprule
\multicolumn{2}{c}{Data set} & \multicolumn{5}{c}{CIFAR-10} & \multicolumn{4}{c}{CIFAR-100}\\    
\midrule
\multicolumn{2}{c}{Noise type} & \multicolumn{4}{c}{sym.} & \multicolumn{2}{c}{asym.} &  \multicolumn{4}{c}{sym.} \\
\midrule

Method/ noise ratio & &  20\% & 50\% & 80\% & 90\% & 40\%& 49\% & 20\% & 50\% & 80\% & 90\% \\
\midrule

\multirow{2}{*}{DivideMix~\cite{li2020dividemix} }& Best & 96.22 & 94.93 & 93.33 & 76.49& 93.24 & 82.90 & 78.03 & 74.87 & \textbf{62.74} & 29.79 \\
  & Last & 96.01 & 94.68 & 92.99 & 75.45& 91.79 & 75.57 & 77.43 & 74.23 & \textbf{62.01} & 29.37\\
  & $|\mathcal{X'}|/|\mathcal{X}|$ & 1.0 & 1.0 & 1.0 & 1.0 & 1.0 & 1.0 & 1.0 & 1.0 & 1.0& 1.0\\
  & $|\mathcal{U'}|/|\mathcal{U}|$ & 4.0 & 1.0 & 0.25 & 0.11 & 1.5 & 1.04 & 4.0 & 1.0 & 0.25& 0.11\\
\midrule
\multirow{2}{*}{LongMix [ours]   }& Best & \textbf{96.42} & \textbf{96.03} & 84.81 & 73.13&  93.15 & \textbf{87.72} & \textbf{78.61} & 75.35 & 53.7   & 26.78  \\

(overs. clean)  & Last & \textbf{96.18} & \textbf{95.67} & 84.03 & 65.28& 91.35 & \textbf{81.5} & \textbf{78.1} & 74.76 & 53.07 & 26.36 \\
   & $|\mathcal{X'}|/|\mathcal{X}|$ & 1.25 & 2.0 & 5.0 & 10 & 1.6 & 1.96 & 1.25 & 2.0 & 5.0& 10\\
  & $|\mathcal{U'}|/|\mathcal{U}|$ &  4.0 & 1.0 & 0.25 & 0.11 & 1.5 & 1.04 & 4.0 & 1.0 & 0.25& 0.11\\
\midrule
\multirow{2}{*}{LongMix [ours]   }& Best & 96.18 & 95.19 & \textbf{94.09} & \textbf{85.33} & \textbf{93.38} & 83.23 & 78.03 & \textbf{75.84} & 62.24 & \textbf{33.54}\\
 (overs. clean + noisy)   & Last & 95.98 & 94.79 & \textbf{93.73} & \textbf{84.71} &  \textbf{91.87} & 77.18 & 77.56 & \textbf{74.87} & 61.60 & \textbf{33.00}\\
     & $|\mathcal{X'}|/|\mathcal{X}|$ & 1.25 & 2.0 & 5.0 & 10 & 1.6 & 1.96 & 1.25 & 2.0 & 5.0& 10\\
  & $|\mathcal{U'}|/|\mathcal{U}|$ & 5.0 & 2.0 & 1.25 & 1.11 & 2.5 & 2.04 & 5.0 & 2.0 & 1.25& 1.11\\
\bottomrule \\
\end{tabular}
}
\caption{Comparison of the test accuracy between LongMix  with additional MixUp operations between oversampled clean samples (where size of $\mathcal{X}'$  is $|\mathcal{D}|$) , LongMix with additional MixUp operations between oversampled clean and noisy samples (where size of $\mathcal{X}'$ and $\mathcal{U}'$ in~\eqref{eq:X_prime_U_prime} are $|\mathcal{D}|$), and the baseline in~\cite{li2020dividemix}  (with sizes of $\mathcal{X}'$ and $\mathcal{U}'$ being $|\mathcal{X}|$)  , using the same number of iterations on CIFAR-10 and CIFAR-100 under symmetric (ranging from 20\% to 90\%) and asymmetric (ranging from 40\% and 49\%) noise.}
\label{tab:longmix_sameiter}
\end{table}

\begin{figure}[t]
\centering
\includegraphics[width=0.6\columnwidth]{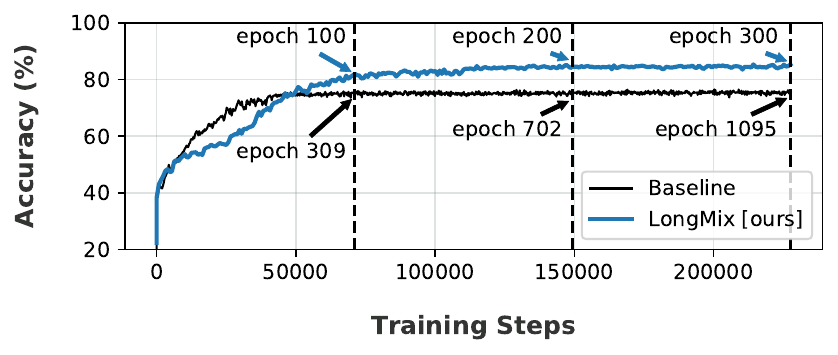}
\caption{Test accuracy versus number of training steps (or iterations) for CIFAR-10 at 90\% symmetric noise for our proposed LongMix (where sizes of $\mathcal{X}'$ and $\mathcal{U}'$ in~\eqref{eq:X_prime_U_prime} are $|\mathcal{D}|$) and the baseline (with sizes of $\mathcal{X}'$ and $\mathcal{U}'$ being $|\mathcal{X}|$~\cite{li2020dividemix}).
}
\label{fig:acc_iters}
\end{figure}

\begin{table}[ht]
\centering
\footnotesize
\scalebox{0.6}{
\begin{tabular}{cc|cccc|cc||cccc| c}
\toprule
\multicolumn{2}{c}{Data set} & \multicolumn{5}{c}{CIFAR-10} & \multicolumn{5}{c}{CIFAR-100}\\    
\midrule
\multicolumn{2}{c}{Noise type} & \multicolumn{4}{c}{sym.} & \multicolumn{2}{c}{asym.} &  \multicolumn{4}{c}{sym.}  & asym.\\
\midrule

Method/ noise ratio & &  20\% & 50\% & 80\% & 90\% & 40\%& 49\% & 20\% & 50\% & 80\% & 90\% & 40\% \\
\midrule
\multirow{2}{*}{Cross-Entropy \cite{li2020dividemix}}& Best & 86.8 & 79.4 & 62.9 & 42.7 & 85.0 & - & 62.0 & 46.7 & 19.9 & 10.1 & -\\
    & Last & 82.7 & 57.9 & 26.1 & 16.8 & 72.3 & - & 61.8 & 37.3 & 8.8 & 3.5& -\\
\midrule
\multirow{2}{*}{Coteaching+ \cite{yu2019does}} & Best & 89.5 & 85.7 & 67.4& 47.9& - & -&65.6& 51.8 & 27.9 & 13.7& -\\
    & Last & 88.2 & 84.1 & 45.5& 30.1& - & -&64.1& 45.3 & 15.5 & 8.8& -\\
\midrule
\multirow{2}{*}{MixUp \cite{zhang2017mixup}} &Best & 95.6 & 87.1 & 71.6& 52.2& - & - & 67.8& 57.3 & 30.8 & 14.6& -\\
 &Last & 92.3 & 77.3 & 46.7& 43.9& - & - & 66.0& 46.6 & 17.6 & 8.1& -\\
\midrule
\multirow{2}{*}{Meta-Learning \cite{li2019learning}}&Best & 92.9 & 89.3 & 77.4& 58.7& 89.2 & - & 68.5& 59.2 & 42.4 & 19.5& - \\
   &Last & 92.0 & 88.8 & 76.1& 58.3& 88.6 & - & 67.7& 58.0 & 40.1 & 14.3& - \\
\midrule
\multirow{2}{*}{M-correction \cite{arazo2019unsupervised}}&Best& 94.0 & 92.0 & 86.8& 69.1& 87.4 & - & 73.9& 66.1 & 48.2 & 24.3& - \\
   &Last& 93.8 & 91.9 & 86.6& 68.7& 86.3 & - & 73.4& 65.4 & 47.6 & 20.5& - \\
\midrule
\multirow{2}{*}{DivideMix \cite{li2020dividemix}}& Best &96.1 & 94.6 & 93.2 & 76.0& 93.4 & 83.7 & 77.3 & 74.6 & 60.2 & 31.5& 59.1 \\
  & Last & 95.7 & 94.4 & 92.9 & 75.4& 92.1 & 76.3 & 76.9 & 74.2 & 59.6 & 31.0& 53.5 \\
\midrule
\multirow{2}{*}{ELR+~\cite{liu2020early}}& Best & 95.8 & 94.8 & 93.3 & 78.7 & 93.0 & - & 77.6 & 73.6 & 60.8 & 33.4& - \\
  & Last & - & - & - & -& - & - & - & - & - & - & -\\
\midrule
\multirow{2}{*}{LongReMix [ours] }& Best & \textbf{96.3$\pm$0.1} & \textbf{95.1$\pm$0.1} & \textbf{93.8$\pm$0.2} & \textbf{79.9$\pm$2.7} & \textbf{94.7$\pm$0.1} & \textbf{84.4$\pm$0.8} & \textbf{77.9$\pm$0.2} & \textbf{75.5$\pm$0.2} & \textbf{62.3$\pm$0.5} & \textbf{34.7$\pm$0.3}& \textbf{59.8$\pm$0.1}\\
    & Last & \textbf{96.0$\pm$0.1} & \textbf{94.8$\pm$0.1} & \textbf{93.3$\pm$0.2} & \textbf{79.1$\pm$3.1} & \textbf{94.3$\pm$0.1} & \textbf{77.8$\pm$0.4} & \textbf{77.5$\pm$0.2} & \textbf{74.9$\pm$0.2} & \textbf{61.7$\pm$0.5} & \textbf{30.7$\pm$5.9}& \textbf{54.9$\pm$0.4}\\
\bottomrule \\
\end{tabular}
}
\caption{Results using PRN18 on CIFAR-10 and CIFAR-100 under symmetric (ranging from 20\% to 90\% and asymmetric (ranging from 40\% and 49\%) noises. Results from related approaches are as presented in~\cite{li2020dividemix}.} 
\label{tab:results_cifar}
\end{table}

\subsection{Testing Hypothesis One: Improving the Precision in Differentiating Clean and Noisy Samples}
\label{sec:hypothesis_one_empirical_results}

In this section, we show empirically that our approach to differentiate between clean and noisy samples leads to higher precision and lower recall than the method based on the small-loss strategy~\cite{li2020dividemix}, which in turn enables higher classification accuracy.
For the experiments below, we measure the $\text{Precision} = \frac{\mathrm{TP}}{\mathrm{TP}+\mathrm{FP}}$ and $\text{Recall} = \frac{\mathrm{TP}}{\mathrm{TP}+\mathrm{FN}}$ results after the first training stage (HCT) and the classification accuracy after the second stage of training (i.e., guided training), where $\mathrm{TP}$ refers to the samples correctly predicted as clean,  $\mathrm{FP}$ denotes the noisy samples incorrectly predicted as clean, and $\mathrm{FN}$ denotes the clean samples incorrectly predicted as noisy.
We first show in Figure~\ref{fig:prec_acc} that in general, higher accuracy is correlated with a higher precision of the predicted clean set, particularly for high noise rates, where a clean set precision increase is obtained by varying the confidence window $\zeta \in \{1,3,5,10\}$. Increasing the value of $\zeta$ leads to a higher precision, which is related to a high accuracy, as shown in Figure~\ref{fig:prec_acc} . Table~\ref{tab:confidence_window} displays the results of LongReMix using different values of $\zeta \in \{1,3,5,10\}$, for CIFAR-10, under symmetric (ranging from 80\% to 90\%) and asymmetric (ranging from 40\% and 49\%) noises. Note that in general, precision increases and recall decreases with larger $\zeta$ values, and classification accuracy reaches a peak at around $\zeta=5$. Hereafter, we fixed the confidence window at $\zeta=5$ in~\eqref{eq:redefine_clean_noisy_sets}. Figure~\ref{fig:loss_epoch} (a) shows how the training loss can be unstable during training for high levels of noise rate, which harms the precision of the predicted clean set. By increasing $\zeta$, we can improve the precision of clean set, as displayed in Figure~\ref{fig:loss_epoch} (b).


We evaluate the precision and recall of the clean set $\mathcal{X}_1^{(e)}$ from~\eqref{eq:redefine_clean_noisy_sets} in the last epoch of the first stage of training (HCT), 
compared to the clean set $\mathcal{X}$ from~\eqref{eq:set_U_X} that relies on the small loss result from the last epoch (DivideMix Baseline~\cite{li2020dividemix}). 
We assess that by computing Precision and Recall 
of the sets $\mathcal{X}_1^{(e)}$ from~\eqref{eq:redefine_clean_noisy_sets} (HCT [ours]) and $\mathcal{X}$ from~\eqref{eq:set_U_X} (DivideMix Baseline).
Figure~\ref{fig:precison}-(a) shows the Precision vs Recall of predicted clean set for CIFAR-10 with 40\% asymmetric noise, where results are obtained by varying the threshold $\tau$ applied to  $p(\text{clean}|\ell,\gamma)$ to form $\mathcal{X}_1^{(e)}$ and $\mathcal{X}$. 
We highlight the value of $\tau=0.5$, which is the default value~\cite{li2020dividemix} that we use to split the clean and noisy samples. Notice that in this highly asymmetric noise scenario, the curve from HCT shows a better trade-off than the Baseline.
Figure~\ref{fig:precison}-(b,c) shows that $\mathcal{X}_1^{(e)}$ from HCT trades off a higher precision for a lower recall, compared with $\mathcal{X}$ from the Baseline for several types of noise,  As shown below, this has a large influence on the training efficacy of LongReMix. 
The only exception to this pattern in Figure~\ref{fig:precison}-(b,c) is the 90\% symmetric case for CIFAR-100, where precision is smaller and recall larger for HCT, compared with the baseline.  We notice that for this particular case with high-noise rate and large number of classes, the initial clean set has a very low clean sample classification accuracy (around 13\%) at the start of the training, which contains a fair amount of noise which can be quickly overfit, damaging the rest of the training.  This issue can be fixed by increasing the confidence window $\zeta$ from 5 to 10, when we notice that the usual pattern of higher precision and lower recall for HCT compared with the baseline is restored. According to the  results in Figure~\ref{fig:precison}, the use of this high-confidence set can improve precision by as much as 30\%, with an average recall reduction of around of 10\%.
Also, Table~\ref{tab:ablation} shows that the high-confidence clean set from HCT can improve the classification results up to by around 2\%, compared with the DivideMix baseline~\cite{li2020dividemix}.

\subsection{Testing Hypothesis Two: Oversampling the Clean Data to Increase the Robustness of EVR}
\label{sec:hypothesis_two_empirical_results}

Figure~\ref{fig:acc_iters} shows the test accuracy versus the number of training steps (iterations) for LongMix compared to the baseline~\cite{li2020dividemix}, for CIFAR-10 at 90\% symmetric noise. This figure shows that adding more MixUp iterations per epoch, as in LongMix, is not equivalent to adding more epochs, as in baseline~\cite{li2020dividemix}.
This shows evidence for the claim in Section~\ref{sec:guided_training} that a simple increase in the number of epochs is not equivalent to adding more MixUp iterations, as we propose for LongMix.
Table~\ref{tab:longmix_sameiter} shows further evidence for this claim by comparing LongMix and DivideMix baseline~\cite{li2020dividemix} using the same number of training iterations for different noise rates on CIFAR-10 and CIFAR-100. In Table~\ref{tab:longmix_sameiter}, we also evaluate the use of LongMix with two oversampling approaches combined with MixUp. The first approach (labelled as 'LongMix [ours] (overs. clean)') uses LongMix with an oversampling of the clean set, followed by MixUp operations among the clean samples. The second approach (labelled as 'LongMix [ours] (overs. clean + noisy)') oversamples the clean and noisy samples and promotes additional MixUp operations between these two sets. The second approach shows more stable results for the studied noise rates and it is the one selected to be used in the following experiments in this work. We can also see in Table~\ref{tab:longmix_sameiter} that LongMix is in general more accurate than standard DivideMix for most cases, with improvements of up to 9\%.

\subsection{Comparison with the State-of-the-Art}

\begin{table}[ht]
\centering
\footnotesize
\scalebox{1.0}{
\begin{tabular}{@{}p{2.7cm}@{}p{0.5cm}p{0.5cm}p{0.6cm}p{0.5cm}p{0.5cm}p{0.5cm}}
\toprule
Data set & \multicolumn{3}{c}{CIFAR-10} & \multicolumn{3}{c}{CIFAR-100}\\    
\midrule

Method/ noise ratio & DN (32\%) & RN (38\%) & VGG (34\%) & DN (34\%) & RN (37\%) & VGG (37\%) \\
\midrule
CE + RoG &  68.33 & 64.15 & 70.04 & 61.14 & 53.09 & 53.64\\
Bootstrap + RoG &  68.38 & 64.03 & 70.11 & 54.71 & 53.30 & 53.76\\
Forward + RoG &  68.20 & 64.24 & 70.09 & 53.91 & 53.36 & 53.63\\
Backward + RoG &  68.66 & 63.45 & 70.18 & 54.01 & 53.03 & 53.50\\
D2L + RoG &  68.57 & 60.25 & 59.94 & 31.67 & 39.92 & 45.42\\
DivideMix*  &  84.57 & 81.61 & 85.71 & 68.40 & 66.28 & 66.84\\
LongReMix [ours] &  \textbf{85.13} & \textbf{82.51} & \textbf{85.90} & \textbf{69.03} & \textbf{66.70} & \textbf{67.42}\\
\bottomrule \\
\end{tabular}
}
\caption{Results for Instanced-based Noise. Results from baseline methods are as presented in~\cite{rog}. Methods marked by * denote re-implementations based on public code.} 
\label{tab:res_semantic}
\end{table}

\begin{table}
\centering
\scalebox{0.8}{
\begin{tabular}{lcccc}
\toprule
Method/ noise ratio & 20\% & 40\% & 60\%& 80\% \\
\midrule
 Cross-entropy~\cite{FaMUS}    & 47.36 & 42.70 & 37.30 & 29.76 \\
 MixUp~\cite{zhang2017mixup}          & 49.10 & 46.40 & 40.58 & 33.58 \\
 DivideMix~\cite{li2020dividemix}     & 50.96 & 46.72 & 43.14 & 34.50 \\
 MentorMix~\cite{jiang2020beyond}  & 51.02 & 47.14 & 43.80 & 33.46 \\
 FaMUS~\cite{FaMUS}  & 51.42 & 48.06 & 45.10 & 35.50 \\
 \textbf{LongReMix (Ours)} & \textbf{56.03$\pm$0.5} & \textbf{50.69$\pm$0.3} & \textbf{46.81$\pm$0.3} & \textbf{38.24$\pm$0.2} \\
\bottomrule \\
\end{tabular}
}
\caption{Test accuracy (\%) for Red Mini-ImageNet~\cite{jiang2020beyond}. Results from baseline methods are as presented in \cite{FaMUS}. Top methods are in \textbf{bold}.}
\label{tab:results_red_noise}
\end{table}

\begin{table}[!ht]
\footnotesize
\centering
\scalebox{1.0}{
\begin{tabular}{lcc}
\toprule
Method  & Top 1 & Top 5  \\
\midrule
 Decoupling~\cite{malach2017decoupling}     & 62.54 & 84.74   \\
 D2L~\cite{ma2018dimensionality}            & 62.68 & 84.00   \\
 MentorNet~\cite{jiang2018mentornet}      & 63.00 & 81.40   \\
 Co-teaching~\cite{han2018co}    & 63.58 & 85.20  \\
 Iterative-CV~\cite{chen2019understanding}   & 65.24 & 85.34  \\
 DivideMix~\cite{li2020dividemix}      & 77.32 & 91.64  \\
 ELR+~\cite{liu2020early}      & 77.78 & 91.68  \\
 LongReMix~[ours]     & \textbf{78.92}  & \textbf{92.32} \\
 \bottomrule \\
 
\end{tabular}
}
\caption{Results for WebVision~\cite{li2017WebVision}. Results from baseline methods are as presented in \cite{li2020dividemix}.}
\label{tab:res_WebVision}
\end{table}

\begin{table}[t]
\footnotesize
\centering
\scalebox{1.0}{
\begin{tabular}{lcc}
\toprule
Method & Test Accuracy \\
\midrule
 Cross-Entropy~\cite{li2020dividemix}  & 69.21  \\
 M-correction \cite{arazo2019unsupervised}   & 71.00 \\
 PENCIL\cite{yi2019probabilistic} &   73.49 \\
 DeepSelf~\cite{han2019deep} & 74.45 \\
 CleanNet~\cite{lee2018cleannet}    & 74.69 \\
 DivideMix~\cite{li2020dividemix}      & \textbf{74.76} \\
 LongReMix~$\dagger$~[ours]      &  74.38 \\

 \bottomrule
 \\
\end{tabular}
}
\caption{Results for Clothing1M~\cite{xiao2015learning}. Results from baseline methods are as presented in \cite{li2020dividemix}. The marker $\dagger$ denotes the model is trained from scratch.}
\label{tab:sota_cl}
\end{table}

\begin{table}[t]
\footnotesize
\centering
\scalebox{1.0}{
\begin{tabular}{lcc}
\toprule
Method  & from pre-trained & from scratch \\
\midrule
 Cross-Entropy   & 81.44& -  \\
 CleanNet   & 83.95 & - \\
 DeepSelf  & 85.10 & - \\
 DivideMix*  & 86.91 &  75.53 \\
 LongReMix [ours]  & \textbf{87.39} &  \textbf{78.57} \\
 \bottomrule
 \\
\end{tabular}
}
\caption{Results for Food-101N~\cite{lee2018cleannet}. Methods marked by * denote re-implementations based on public code.}
\label{tab:res_food}
\end{table}

All comparisons in this section are performed with the same network architecture and trained for the same number of epochs as the compared methods.
For CIFAR-10 and CIFAR-100, we evaluate our model using different levels of symmetric label noise ranging from 20\% to 90\%. We also consider asymmetric noise, with noise rates of 40\% and 49\%. 
For all results, we train the model three times, using different initialisation, and report the results with the mean $\pm$ standard deviation.
We report both the best test accuracy across all epochs and the averaged test accuracy over the last 10 epochs of training, similar to~\cite{li2020dividemix}. Table~\ref{tab:results_cifar} shows that for CIFAR-10 and CIFAR-100 data sets, our method obtains better results for all evaluated noise rates. LongReMix displays a higher improvement for large symmetric noise and asymmetric noise scenarios, which can be considered as the most challenging cases. We believe that the improvement over higher noise rates is due to the LongMix approach, which runs a large number of MixUp operations proportional to the size of the training set. The retraining with high confidence samples also improves the results for asymmetric noise. 
Using the mean $\pm$ standard deviation results on Table~\ref{tab:results_cifar}, 
we notice that LongReMix is in general at least 3 standard deviations from the competing methods, which can be considered a significant result (recall that 99.7\% of samples following a normal distribution lie within 3 standard deviations of the mean).
The results for instanced-based noise~\cite{rog} in
Table~\ref{tab:res_semantic} shows again the superiority of our approach compared to the related work.
We also show results on the Red Mini-ImageNet problem from the CNWL dataset~\cite{jiang2020beyond,FaMUS} on Table~\ref{tab:results_red_noise}. Note that in this table, we also train the model three times, using different initialisations, and report the results with the mean $\pm$ standard deviation.
This table shows again that LongReMix has significantly better accuracy than competing SOTA approaches (at least three standard deviations away from the SOTA).

Also, we evaluate our method on large-scale data sets. For WebVision, 
Table~\ref{tab:res_WebVision} shows the Top-1 and Top-5 accuracy, where LongReMix displays better results than competing methods. 
For the Clothing1M evaluation, 
the competing methods rely on a pre-trained ImageNet model for training on Clothing1M. In our experiments, we did not observe any improvement with pre-trained models, and therefore we trained from scratch with 128k images from Clothing1M. The results in Table~\ref{tab:sota_cl} show that our model, trained from scratch and with a reduced training set, obtained comparable results to the competing approaches. Lastly, Table~\ref{tab:res_food} summarizes the results for Food-101N. 
For this problem, we evaluate our approach with a pre-trained model and trained from scratch, and LongReMix outperforms all other approaches in both scenarios.

\subsection{Statistical Analysis}

\begin{figure}[!t]
\centering
\includegraphics[width=0.8\columnwidth]{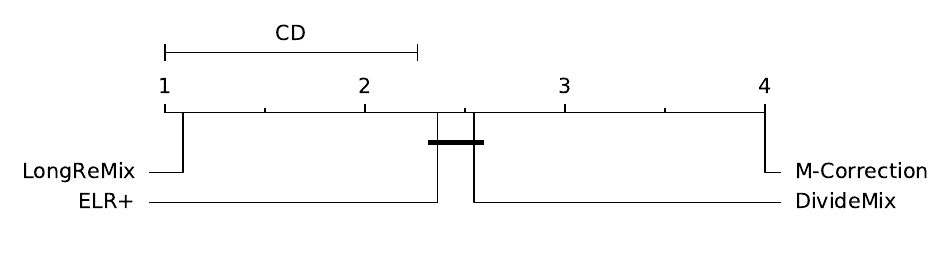}
\caption{Comparison of  LongReMix, ELR+~\cite{liu2020early}, DivideMix~\cite{li2020dividemix} and M-correcction~\cite{arazo2019unsupervised} using the Friedman-Nemenyi statistical test, which produced a critical distance (CD) value of 1.26. 
This CD value is used to estimate if two methods are  different at level significance level $0.1$. 
Groups of classifiers that are not significantly different are connected with a thick horizontal line (e.g., ELR+ and DivideMix are not significantly different).}
\label{fig:critical_distance}
\end{figure}

We compare our LongReMix 
with ELR+~\cite{liu2020early}, DivideMix~\cite{li2020dividemix} and M-correcction~\cite{arazo2019unsupervised}, relying on their best results over multiple data sets, and using the statistical test proposed by Dem{\v{s}}ar~\cite{demvsar2006statistical}.
Specifically, we use the Friedman non-parametric test, with significance level $0.1$, 
where the null hypothesis indicates that
all methods perform equally well.
For this analysis, we use CIFAR-10, CIFAR-100 and WebVision, comprising a total of 10 different independent experiments (including the different noise rates) for each method. 
The Friedman test produces a $\text{p-value}=3.47 \times 10^{-3}$,
which rejects the null hypothesis, suggesting that at least one of the methods is statistically different from the others. 
Thus, to identify the groups of methods that present statistical similarity in multiple comparisons, we apply the post-hoc \textit{Nemenyi} test to obtain the average ranks and calculate the Critical Difference (CD) value of the results, where the results from two methods are considered to be significantly different if the corresponding average
ranks differ by at least the CD value. The test produced a CD value of 1.26, which means that a rank distance above 1.26 represents significantly different methods at $0.1$ confidence level. 
Figure~\ref{fig:critical_distance} shows the comparison of these results through the CD diagram, where we show that LongReMix is significantly better than all other methods, with ELR+ and DivideMix not being statistically different from each other.

\subsection{Ablation Study}

\begin{table*}[t!]
\centering
\footnotesize
\scalebox{0.7}{
\begin{tabular}{@{}p{1.2cm}p{0cm}|p{0.5cm}p{0.5cm}p{0.5cm}p{0.6cm}|p{0.5cm}p{0.6cm}||p{0.5cm}p{0.5cm}p{0.5cm}p{0.6cm}||p{0.6cm}p{0.7cm}p{0.7cm} || c@{}}
\toprule
\multicolumn{2}{c}{Data set} & \multicolumn{6}{c}{CIFAR-10} & \multicolumn{4}{c}{CIFAR-100}& Webv.& Cloth. & Food. & Mean Rank\\    
\midrule
\multicolumn{2}{c}{Noise type} & \multicolumn{4}{c}{sym.} & \multicolumn{2}{c}{asym.} &  \multicolumn{4}{c}{sym.} & - & - & - & -  \\
\midrule

\multicolumn{2}{@{}p{2.17cm}}{Method/ n. ratio} &  20\% & 50\% & 80\% & 90\% & 40\%& 49\% & 20\% & 50\% & 80\% & 90\%& -& -& - & -  \\
\midrule
\multirow{2}{*}{LongReMix }& Best & \textit{\textbf{96.25}} & \textbf{95.01} & \textbf{93.88} & \textbf{81.98} & \textit{\textbf{94.64}} & \textbf{84.68} & \textbf{77.82} & \textbf{75.59} & \textit{\textbf{62.92}} & \textit{\textbf{33.80}}& \textit{\textbf{78.92}}&  \textbf{74.38} & \textit{\textbf{87.39}} & 1.46\\
    & Last & \textit{\textbf{96.02}} & \textbf{94.72} & \textbf{93.37} & \textbf{81.35} & \textit{\textbf{94.32}} & 76.08 & \textbf{77.52} & \textbf{75.11} & \textit{\textbf{62.34}} & \textit{\textbf{33.25}} & \textit{\textbf{78.00}} & 73.00 & \textit{\textbf{87.29}} & 1.69\\
\midrule
\multirow{2}{*}{LongMix} & Best & 96.18 & \textit{\textbf{95.19}} & \textit{\textbf{94.09}} & \textit{\textbf{85.33}}& 93.38 & 83.23 & \textit{\textbf{78.03}} & \textit{\textbf{75.84}} & \textbf{62.24} & \textbf{33.54}& \textbf{78.44} & 74.05 & \textbf{87.21} & 1.92\\
    & Last & \textbf{95.98} & \textit{\textbf{94.79}} & \textit{\textbf{93.73}} & \textit{\textbf{84.71}}& 91.87 & \textbf{77.18} & \textit{\textbf{77.56}}& \textit{\textbf{74.87}} & \textbf{61.60} & \textbf{33.00} & 77.72& \textit{\textbf{73.25}} & \textbf{87.12} & 1.69\\
\midrule
\multirow{2}{*}{Retrain} & Best & \textbf{96.23} & 94.85 & 92.86& 78.47& \textbf{94.59} & \textit{\textbf{85.10}} &77.20& 74.41 & 60.29 & 30.61 & 77.84 & \textbf{74.30} & 87.16 & 2.61\\
    & Last & 95.89 & 94.60 & 92.54& 77.51& \textbf{94.31} & \textit{\textbf{80.88}} &76.89& 73.89 & 59.88 & 30.37 & \textbf{77.84} & \textbf{73.21} & 86.98 & 2.61\\

\bottomrule  \\
\end{tabular}
}
\caption{Ablation Study Results. The italic bold, bold, and regular numbers represent respectively the ranking of first, second and third results in accuracy.  Last column shows the average rank of each approach (smaller is better).}
\label{tab:ablation}
\end{table*}

We analyze the effect of the different components of our proposal in an ablation study, shown in Table~\ref{tab:ablation}. 
We first evaluate our approach without LongMix -- this approach is referred to as ``Retrain''. 
Then we evaluate training only with the LongMix, without the second stage of re-training, and the whole model is denoted as LongReMix. 
In general, we can observe that the LongReMix is competitive for all noise scenarios (being best or second best for all cases), but it is generally better for the large-scale data sets.
Considering
different data sets and noise rates, LongReMix shows the best average rank.

\section{Conclusion}

We presented LongReMix, 
a new noisy-label learning algorithm based on an unsupervised learning stage to classify clean and noisy training samples, followed by an SSL stage to minimise the EVR using a labelled set formed by samples classified as clean, and an unlabelled set with samples classified as noisy.


We showed a thorough analysis of LongReMix, providing evidence that our new unsupervised clean sample classification and novel approach to increase the training set size for MixMatch enable a better classification accuracy for models trained with noisy-label samples. 
We also showed that LongReMix reaches state-of-the-art performance on the main noisy-label learning benchmarks of the field, showing robustness to over-fitting in high label noise problems.
Using the statistical test from~\cite{demvsar2006statistical}, we show that the LongReMix results are statistically significant compared to its main competitors DivideMix~\cite{li2020dividemix}, ELR+~\cite{liu2020early}, and M-Correction~\cite{arazo2019unsupervised}.

Although our LongReMix showed state-of-the-art results in the main benchmarks of the field, we plan to investigate further how to improve the precision of the unsupervised classification of clean and noisy samples since this seems to be a critical aspect of these two-stage noisy-label learning algorithms. We will also study new ways to improve the SSL learning accuracy by considering state-of-the-art methods~\cite{chen2020big}. Another important point worth investigating is the role of learning regularisation~\cite{wang2019symmetric,ma2020normalized,wang2019imae} in our framework.
Finally, the study of new types of noise models (e.g., combined closed- and open-set noise~\cite{sachdeva2021evidentialmix}) is also important to consider by future noisy-label learning algorithms.

\section*{Acknowledgments}
The authors would like to thank the support by the Australian Research Council through grants DP180103232, and FT190100525.


\bibliography{egbib}

\begin{thebibliography}{10}
\expandafter\ifx\csname url\endcsname\relax
  \def\url#1{\texttt{#1}}\fi
\expandafter\ifx\csname urlprefix\endcsname\relax\def\urlprefix{URL }\fi
\expandafter\ifx\csname href\endcsname\relax
  \def\href#1#2{#2} \def\path#1{#1}\fi

\bibitem{litjens2017survey}
G.~Litjens, T.~Kooi, B.~E. Bejnordi, A.~A.~A. Setio, F.~Ciompi, M.~Ghafoorian,
  J.~A. Van Der~Laak, B.~Van~Ginneken, C.~I. S{\'a}nchez, A survey on deep
  learning in medical image analysis, Medical image analysis 42 (2017) 60--88.

\bibitem{frenay_survey}
B.~Fr{\'e}nay, M.~Verleysen, Classification in the presence of label noise: a
  survey, IEEE transactions on neural networks and learning systems 25~(5)
  (2013) 845--869.

\bibitem{zhang2016understanding}
C.~Zhang, S.~Bengio, M.~Hardt, B.~Recht, O.~Vinyals, Understanding deep
  learning requires rethinking generalization, in: International Conference on
  Learning Representations (ICLR), 2017.

\bibitem{kim2019nlnl}
Y.~Kim, J.~Yim, J.~Yun, J.~Kim, Nlnl: Negative learning for noisy labels, in:
  Proceedings of the IEEE International Conference on Computer Vision, 2019,
  pp. 101--110.

\bibitem{wang2019symmetric}
Y.~Wang, X.~Ma, Z.~Chen, Y.~Luo, J.~Yi, J.~Bailey, Symmetric cross entropy for
  robust learning with noisy labels, in: Proceedings of the IEEE International
  Conference on Computer Vision, 2019, pp. 322--330.

\bibitem{ren2018learning}
M.~Ren, W.~Zeng, B.~Yang, R.~Urtasun, Learning to reweight examples for robust
  deep learning, in: International Conference on Machine Learning, 2018, pp.
  4334--4343.

\bibitem{nguyen2019self}
T.~Nguyen, C.~Mummadi, T.~Ngo, L.~Beggel, T.~Brox, Self: learning to filter
  noisy labels with self-ensembling, in: International Conference on Learning
  Representations (ICLR), 2020.

\bibitem{li2020dividemix}
J.~Li, R.~Socher, S.~C. Hoi, Dividemix: Learning with noisy labels as
  semi-supervised learning, International Conference on Learning
  Representations (ICLR).

\bibitem{yu2019does}
X.~Yu, B.~Han, J.~Yao, G.~Niu, I.~W. Tsang, M.~Sugiyama, How does disagreement
  help generalization against label corruption?, in: International Conference
  on Machine Learning (ICML), 2019.

\bibitem{arazo2019unsupervised}
E.~Arazo, D.~Ortego, P.~Albert, N.~O’Connor, K.~Mcguinness, Unsupervised
  label noise modeling and loss correction, in: International Conference on
  Machine Learning, 2019, pp. 312--321.

\bibitem{berthelot2019mixmatch}
D.~Berthelot, N.~Carlini, I.~Goodfellow, N.~Papernot, A.~Oliver, C.~A. Raffel,
  Mixmatch: A holistic approach to semi-supervised learning, in: Advances in
  Neural Information Processing Systems (NeurIPS), 2019, pp. 5049--5059.

\bibitem{zhang2017mixup}
H.~Zhang, M.~Cisse, Y.~N. Dauphin, D.~Lopez-Paz, Mixup: Beyond empirical risk
  minimization, in: International Conference on Learning Representations
  (ICLR), 2018.

\bibitem{zhang2018generalization}
C.~Zhang, M.-H. Hsieh, D.~Tao, Generalization bounds for vicinal risk
  minimization principle, arXiv preprint arXiv:1811.04351.

\bibitem{relab}
P.~Albert, D.~Ortego, E.~Arazo, N.~O'Connor, K.~McGuinness, Relab: Reliable
  label bootstrapping for semi-supervised learning, in: 2021 International
  Joint Conference on Neural Networks (IJCNN), IEEE, 2021, pp. 1--8.

\bibitem{area_under_margin}
G.~Pleiss, T.~Zhang, E.~R. Elenberg, K.~Q. Weinberger, Identifying mislabeled
  data using the area under the margin ranking, in: NeurIPS, 2020.

\bibitem{toneva2018empirical}
M.~Toneva, A.~Sordoni, R.~T.~d. Combes, A.~Trischler, Y.~Bengio, G.~J. Gordon,
  An empirical study of example forgetting during deep neural network learning,
  in: International Conference on Learning Representation (ICLR), 2019.

\bibitem{tarvainen2017mean}
A.~Tarvainen, H.~Valpola, Mean teachers are better role models: Weight-averaged
  consistency targets improve semi-supervised deep learning results, in:
  NeurIPS, 2017.

\bibitem{chen2018semi}
Y.~Chen, X.~Zhu, S.~Gong, Semi-supervised deep learning with memory, in:
  Proceedings of the European conference on computer vision (ECCV), 2018, pp.
  268--283.

\bibitem{iscen2019label}
A.~Iscen, G.~Tolias, Y.~Avrithis, O.~Chum, Label propagation for deep
  semi-supervised learning, in: Proceedings of the IEEE/CVF Conference on
  Computer Vision and Pattern Recognition, 2019, pp. 5070--5079.

\bibitem{arazo2020pseudo}
E.~Arazo, D.~Ortego, P.~Albert, N.~E. O’Connor, K.~McGuinness,
  Pseudo-labeling and confirmation bias in deep semi-supervised learning, in:
  2020 International Joint Conference on Neural Networks (IJCNN), IEEE, 2020,
  pp. 1--8.

\bibitem{krizhevsky2009learning}
A.~Krizhevsky, G.~Hinton, et~al., Learning multiple layers of features from
  tiny images, in: Citeseer, 2009.

\bibitem{jiang2020beyond}
L.~Jiang, D.~Huang, M.~Liu, W.~Yang, Beyond synthetic noise: Deep learning on
  controlled noisy labels, ICML, 2020.

\bibitem{li2017WebVision}
W.~Li, L.~Wang, W.~Li, E.~Agustsson, L.~V. Gool, Webvision database: Visual
  learning and understanding from web data., in: CoRR, 2017.

\bibitem{xiao2015learning}
T.~Xiao, T.~Xia, Y.~Yang, C.~Huang, X.~Wang, Learning from massive noisy
  labeled data for image classification, in: Proceedings of the IEEE conference
  on computer vision and pattern recognition, 2015, pp. 2691--2699.

\bibitem{lee2018cleannet}
K.-H. Lee, X.~He, L.~Zhang, L.~Yang, Cleannet: Transfer learning for scalable
  image classifier training with label noise, in: Proceedings of the IEEE
  Conference on Computer Vision and Pattern Recognition, 2018, pp. 5447--5456.

\bibitem{demvsar2006statistical}
J.~Dem{\v{s}}ar, Statistical comparisons of classifiers over multiple data
  sets, The Journal of Machine Learning Research 7 (2006) 1--30.

\bibitem{ma2020normalized}
X.~Ma, H.~Huang, Y.~Wang, S.~Romano, S.~Erfani, J.~Bailey, Normalized loss
  functions for deep learning with noisy labels, in: ICML, 2020.

\bibitem{wang2019imae}
X.~Wang, Y.~Hua, E.~Kodirov, N.~M. Robertson, Imae for noise-robust learning:
  Mean absolute error does not treat examples equally and gradient magnitude's
  variance matters, arXiv preprint arXiv:1903.12141.

\bibitem{jaehwan2019photometric}
L.~Jaehwan, Y.~Donggeun, K.~Hyo-Eun, Photometric transformer networks and label
  adjustment for breast density prediction, in: Proceedings of the IEEE
  International Conference on Computer Vision Workshops, 2019, pp. 0--0.

\bibitem{yuan2018iterative}
B.~Yuan, J.~Chen, W.~Zhang, H.-S. Tai, S.~McMains, Iterative cross learning on
  noisy labels, in: 2018 IEEE Winter Conference on Applications of Computer
  Vision (WACV), IEEE, 2018, pp. 757--765.

\bibitem{han2018pumpout}
B.~Han, G.~Niu, J.~Yao, X.~Yu, M.~Xu, I.~Tsang, M.~Sugiyama, Pumpout: A meta
  approach for robustly training deep neural networks with noisy labels, 2018.

\bibitem{sun2021learning}
H.~Sun, C.~Guo, Q.~Wei, Z.~Han, Y.~Yin, Learning to rectify for robust learning
  with noisy labels, Pattern Recognition (2021) 108467.

\bibitem{miao2015rboost}
Q.~Miao, Y.~Cao, G.~Xia, M.~Gong, J.~Liu, J.~Song, Rboost: Label noise-robust
  boosting algorithm based on a nonconvex loss function and the numerically
  stable base learners, IEEE transactions on neural networks and learning
  systems 27~(11) (2015) 2216--2228.

\bibitem{yu2018learning}
X.~Yu, T.~Liu, M.~Gong, D.~Tao, Learning with biased complementary labels, in:
  Proceedings of the European Conference on Computer Vision (ECCV), 2018, pp.
  68--83.

\bibitem{zhang2020distilling}
Z.~Zhang, H.~Zhang, S.~O. Arik, H.~Lee, T.~Pfister, Distilling effective
  supervision from severe label noise, in: Proceedings of the IEEE/CVF
  Conference on Computer Vision and Pattern Recognition, 2020, pp. 9294--9303.

\bibitem{shu2019meta}
J.~Shu, Q.~Xie, L.~Yi, Q.~Zhao, S.~Zhou, Z.~Xu, D.~Meng, Meta-weight-net:
  Learning an explicit mapping for sample weighting, in: Advances in Neural
  Information Processing Systems, 2019, pp. 1919--1930.

\bibitem{xue2019robust}
C.~Xue, Q.~Dou, X.~Shi, H.~Chen, P.-A. Heng, Robust learning at noisy labeled
  medical images: Applied to skin lesion classification, in: 2019 IEEE 16th
  International Symposium on Biomedical Imaging (ISBI 2019), IEEE, 2019, pp.
  1280--1283.

\bibitem{wang2018iterative}
Y.~Wang, W.~Liu, X.~Ma, J.~Bailey, H.~Zha, L.~Song, S.-T. Xia, Iterative
  learning with open-set noisy labels, in: Proceedings of the IEEE Conference
  on Computer Vision and Pattern Recognition, 2018, pp. 8688--8696.

\bibitem{han2018co}
B.~Han, Q.~Yao, X.~Yu, G.~Niu, M.~Xu, W.~Hu, I.~Tsang, M.~Sugiyama,
  Co-teaching: Robust training of deep neural networks with extremely noisy
  labels, in: Advances in neural information processing systems, 2018, pp.
  8527--8537.

\bibitem{thulasidasan2019combating}
S.~Thulasidasan, T.~Bhattacharya, J.~Bilmes, G.~Chennupati, J.~Mohd-Yusof,
  Combating label noise in deep learning using abstention, in: International
  Conference on Machine Learning, PMLR, 2019, pp. 6234--6243.

\bibitem{sachdeva2021evidentialmix}
R.~Sachdeva, F.~R. Cordeiro, V.~Belagiannis, I.~Reid, G.~Carneiro,
  Evidentialmix: Learning with combined open-set and closed-set noisy labels,
  in: Proceedings of the IEEE/CVF Winter Conference on Applications of Computer
  Vision, 2021, pp. 3607--3615.

\bibitem{patrini2017making}
G.~Patrini, A.~Rozza, A.~Krishna~Menon, R.~Nock, L.~Qu, Making deep neural
  networks robust to label noise: A loss correction approach, in: Proceedings
  of the IEEE Conference on Computer Vision and Pattern Recognition, 2017, pp.
  1944--1952.

\bibitem{rog}
K.~Lee, S.~Yun, K.~Lee, H.~Lee, B.~Li, J.~Shin, Robust inference via generative
  classifiers for handling noisy labels, in: ICML, 2019.

\bibitem{zhang2018generalized}
Z.~Zhang, M.~Sabuncu, Generalized cross entropy loss for training deep neural
  networks with noisy labels, in: Advances in neural information processing
  systems, 2018, pp. 8778--8788.

\bibitem{ding2018semi}
Y.~Ding, L.~Wang, D.~Fan, B.~Gong, A semi-supervised two-stage approach to
  learning from noisy labels, in: 2018 IEEE Winter Conference on Applications
  of Computer Vision (WACV), IEEE, 2018, pp. 1215--1224.

\bibitem{kong2019recycling}
K.~Kong, J.~Lee, Y.~Kwak, M.~Kang, S.~G. Kim, W.-J. Song, Recycling:
  Semi-supervised learning with noisy labels in deep neural networks, IEEE
  Access 7 (2019) 66998--67005.

\bibitem{vgg}
K.~Simonyan, A.~Zisserman, Very deep convolutional networks for large-scale
  image recognition, in: ICLR, 2015.

\bibitem{FaMUS}
Y.~Xu, L.~Zhu, L.~Jiang, Y.~Yang, Faster meta update strategy for noise-robust
  deep learning, in: CVPR, 2021.

\bibitem{han2019deep}
J.~Han, P.~Luo, X.~Wang, Deep self-learning from noisy labels, in: Proceedings
  of the IEEE International Conference on Computer Vision, 2019, pp.
  5138--5147.

\bibitem{yi2019probabilistic}
K.~Yi, J.~Wu, Probabilistic end-to-end noise correction for learning with noisy
  labels, in: Proceedings of the IEEE Conference on Computer Vision and Pattern
  Recognition, 2019, pp. 7017--7025.

\bibitem{deng2009imagenet}
J.~Deng, W.~Dong, R.~Socher, L.-J. Li, K.~Li, L.~Fei-Fei, Imagenet: A
  large-scale hierarchical image database, in: 2009 IEEE conference on computer
  vision and pattern recognition, Ieee, 2009, pp. 248--255.

\bibitem{chen2019understanding}
P.~Chen, B.~Liao, G.~Chen, S.~Zhang, Understanding and utilizing deep neural
  networks trained with noisy labels, in: ICML, 2019.

\bibitem{bossard2014food}
L.~Bossard, M.~Guillaumin, L.~Van~Gool, Food-101--mining discriminative
  components with random forests, in: European conference on computer vision,
  Springer, 2014, pp. 446--461.

\bibitem{he2016identity}
K.~He, X.~Zhang, S.~Ren, J.~Sun, Identity mappings in deep residual networks,
  in: European conference on computer vision, Springer, 2016, pp. 630--645.

\bibitem{szegedy2017inception}
C.~Szegedy, S.~Ioffe, V.~Vanhoucke, A.~Alemi, Inception-v4, inception-resnet
  and the impact of residual connections on learning, in: Proceedings of the
  AAAI Conference on Artificial Intelligence, Vol.~31, 2017.

\bibitem{he2016deep}
K.~He, X.~Zhang, S.~Ren, J.~Sun, Deep residual learning for image recognition,
  in: Proceedings of the IEEE conference on computer vision and pattern
  recognition, 2016, pp. 770--778.

\bibitem{li2019learning}
J.~Li, Y.~Wong, Q.~Zhao, M.~S. Kankanhalli, Learning to learn from noisy
  labeled data, in: Proceedings of the IEEE Conference on Computer Vision and
  Pattern Recognition, 2019, pp. 5051--5059.

\bibitem{liu2020early}
S.~Liu, J.~Niles-Weed, N.~Razavian, C.~Fernandez-Granda, Early-learning
  regularization prevents memorization of noisy labels, in: NeurIPS, 2020.

\bibitem{malach2017decoupling}
E.~Malach, S.~Shalev-Shwartz, Decoupling" when to update" from" how to update",
  in: Advances in Neural Information Processing Systems, 2017, pp. 960--970.

\bibitem{ma2018dimensionality}
X.~Ma, Y.~Wang, M.~E. Houle, S.~Zhou, S.~Erfani, S.~Xia, S.~Wijewickrema,
  J.~Bailey, Dimensionality-driven learning with noisy labels, in:
  International Conference on Machine Learning, 2018, pp. 3355--3364.

\bibitem{jiang2018mentornet}
L.~Jiang, Z.~Zhou, T.~Leung, L.-J. Li, L.~Fei-Fei, Mentornet: Learning
  data-driven curriculum for very deep neural networks on corrupted labels, in:
  International Conference on Machine Learning, 2018, pp. 2304--2313.

\bibitem{chen2020big}
T.~Chen, S.~Kornblith, K.~Swersky, M.~Norouzi, G.~Hinton, Big self-supervised
  models are strong semi-supervised learners, in: NeurIPS, 2020.

\end{thebibliography}

\newpage

\begin{minipage}{0.3\textwidth}
\begin{tikzpicture}[remember picture, overlay]
\node[text width=\linewidth]{
\begin{figure}[H]\centering
    \vspace{-2cm}
    \includegraphics[height=18cm, width=0.7\linewidth, keepaspectratio]{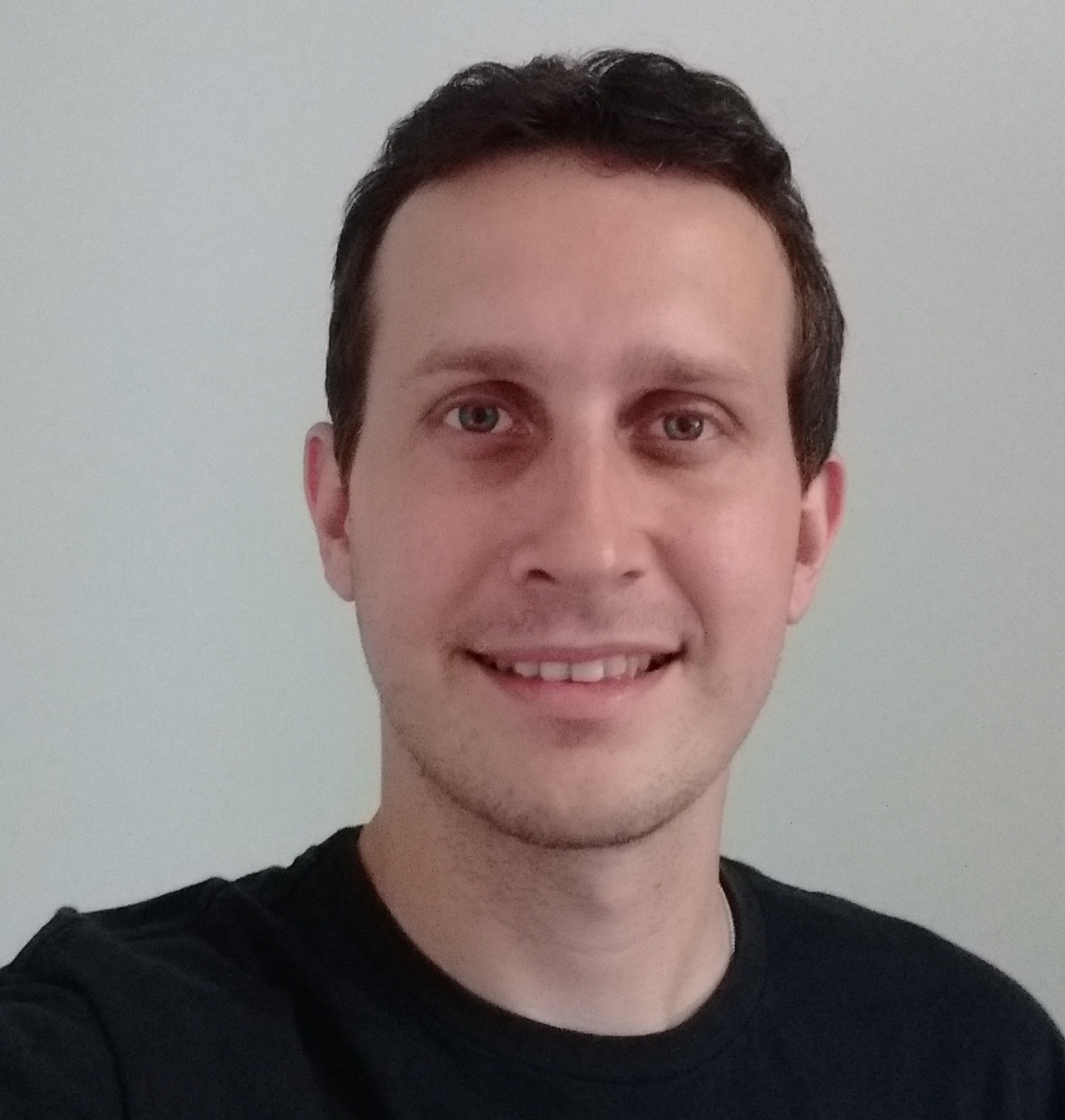}\\
    \vspace{1cm}
    \includegraphics[height=18cm, width=0.7\linewidth, keepaspectratio]{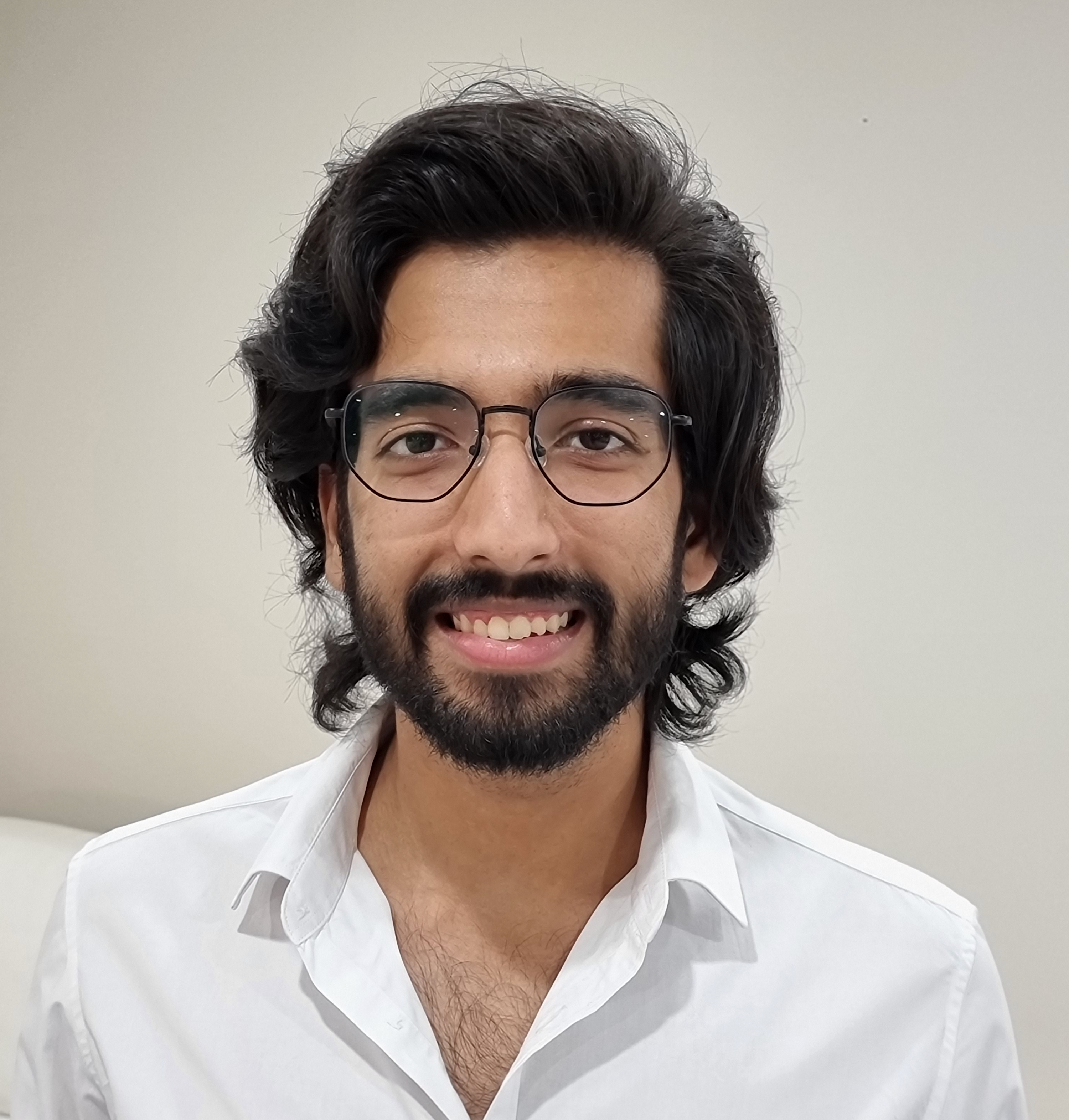}\\
    \vspace{1cm}
    \includegraphics[height=18cm, width=0.7\linewidth, keepaspectratio]{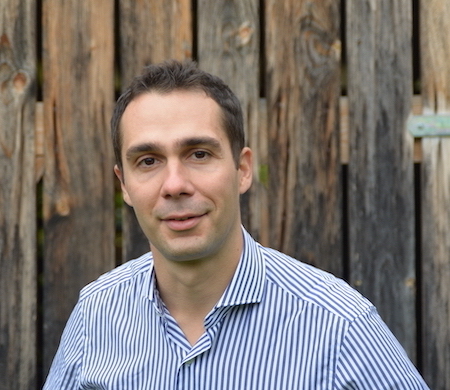}\\
    \vspace{2cm}
    \includegraphics[height=18cm, width=0.7\linewidth, keepaspectratio]{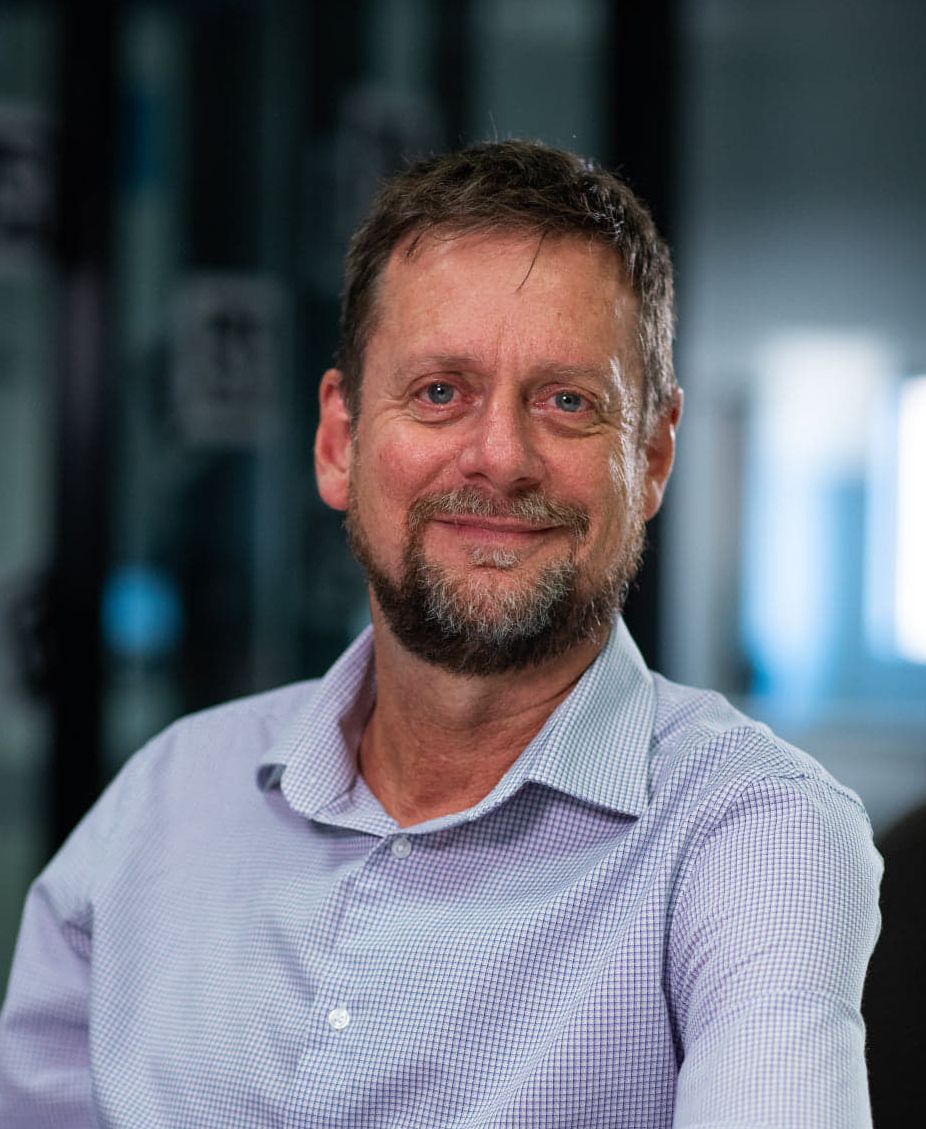}\\
    \vspace{1cm}
    \includegraphics[height=18cm, width=0.7\linewidth, keepaspectratio]{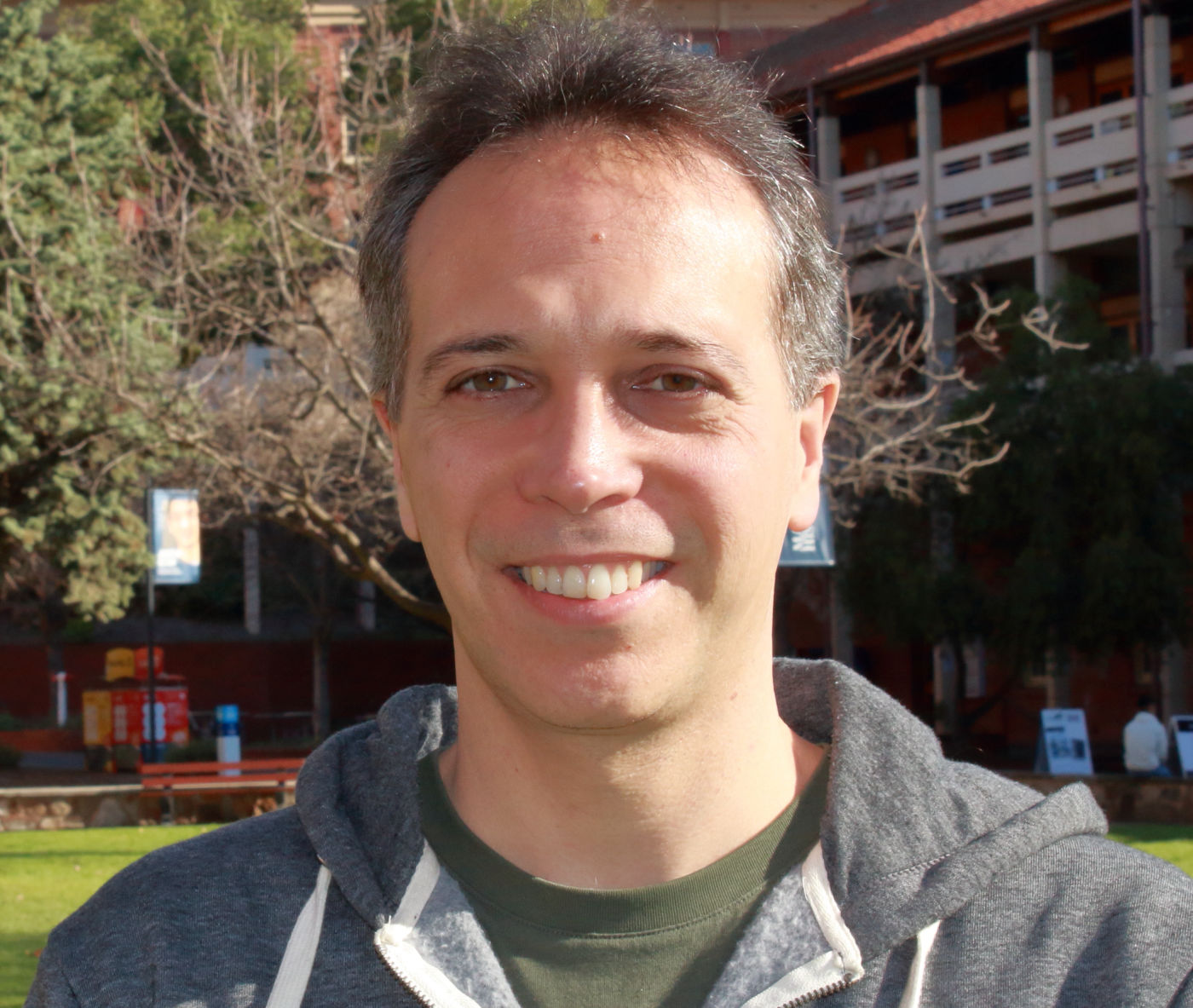}
\end{figure}
};
\end{tikzpicture}
\end{minipage}%
\hfill
\begin{minipage}{0.60\textwidth}\fontsize{8pt}{12pt}\selectfont
\begin{tabular}{p{\textwidth}}
\thispagestyle{empty}
Filipe R. Cordeiro is a professor in the Department of Computing at Universidade Federal Rural de Pernambuco (UFRPE). In 2015, he received his Ph.D. in computer science from the Federal University of Pernambuco (UFPE). Filipe’s mains contributions are in the area of computer vision, medical image analysis, and machine learning.
\\[1\baselineskip]
Ragav Sachdeva is a Ph.D. student in the Visual Geometry Group at the University of Oxford, supervised by Prof. Andrew Zisserman. He obtained his undergraduate degree in computer science at the University of Adelaide, where he did his honours thesis with Prof. Gustavo Carneiro.
\\[1\baselineskip]
Vasileios Belagiannis is a professor in the Faculty of Computer Science at Otto von Guericke University Magdeburg. His research deals with topics such as representation learning, uncertainty estimation, multi-modal learning, learning with different forms of supervision, learning algorithm for noisy labels, few-shot learning and meta-learning.
\\[1\baselineskip]
Ian Reid is the Head of the School of Computer Science at the University of Adelaide, and the senior researcher at the Australian Institute for Machine Learning. His research interests include robotic and active vision, visual tracking, SLAM, human motion capture and intelligent visual surveillance.
\\[1\baselineskip]
Gustavo Carneiro is a professor in the School of Computer Science at the University of Adelaide, Director of Medical Machine Learning at the Australian Institute of Machine Learning and an Australian Research Council Future Fellow. His main research interests are in computer vision, medical image analysis and machine learning. He is moving to the CVSSP at the University of Surrey in December 2022.

\end{tabular}
\end{minipage}%

\end{document}